\documentclass[journal]{IEEEtran}
\ifCLASSINFOpdf
\else
\fi

\usepackage{amsmath}
\usepackage{algorithmic}
\usepackage{array}
\usepackage{booktabs}

\usepackage{color}
\newcommand{\red}[1]{\textcolor{black}{#1}} 
\newcommand{\rev}[1]{\textcolor{black}{#1}} 
\newcommand{\abk}[1]{\textcolor{black}{#1}} 

\usepackage{graphicx}%
\usepackage{amsmath,amssymb,amsfonts}%
\usepackage{amsthm}%
\usepackage{mathrsfs}%
\usepackage{xcolor}%
\usepackage{textcomp}%
\usepackage{manyfoot}%
\usepackage{booktabs}%
\usepackage{multirow}%
\usepackage{algorithm}%
\usepackage{algorithmic}%
\usepackage{listings}%
\usepackage{adjustbox}
\usepackage{wrapfig}
\usepackage{pifont}
\usepackage[T1]{fontenc}
\usepackage[inline]{enumitem}
\usepackage{amssymb}
\usepackage{graphicx}

\begin{document}
\title{Evolutionary Computation: A Paradigm of Natural Generative AI and Creative Synthesis}
\author{Abhishek~Gupta$^{\dagger *}$,~\IEEEmembership{Senior Member,~IEEE},
        Yaxin~Shi$^{\dagger}$,
        Ying~Wu,
        Melvin~Wong,
        Ivor~Tsang,~\IEEEmembership{Fellow,~IEEE}
        Thiago~Rios, 
        Stefan~Menzel,
        Bernhard Sendhoff,~\IEEEmembership{Fellow,~IEEE},
        Yaqing~Hou
        and Yew-Soon~Ong$^{*}$,~\IEEEmembership{Fellow,~IEEE}

\thanks{$^{\dagger}$ These authors contributed equally to this work.}%
\thanks{$^{*}$ Corresponding authors: Abhishek Gupta (abhishekgupta@iitgoa.ac.in) and Yew-Soon Ong (ong\_yew\_soon@a-star.edu.sg).}

\thanks{Abhishek Gupta is with the School of Mechanical Sciences, Indian Institute of Technology (IIT) Goa, India (E-mail: abhishekgupta@iitgoa.ac.in)}

\thanks{Yaxin Shi, Ivor Tsang are with Centre for Frontier AI Research (CFAR), Agency for Science, Technology and Research (A*STAR), Singapore (E-mail: shi\_yaxin@a-star.edu.sg, ivor\_tsang@a-star.edu.sg)}

\thanks{Ying Wu and Yaqing Hou are with the School of Computer Science and Technology, Dalian University of Technology (DLUT), China 116024 (E-mails: wwwuyingying@gmail.com, houyq@dlut.edu.cn).}

\thanks{Melvin Wong is with College of Computing and Data Science (CCDS), Nanyang Technological University (NTU), Singapore (E-mail:wong1357@ntu.edu.sg)}

\thanks{Thiago Rios, Stefan Menzel, and Bernhard Sendhoff are with Honda Research Institute Europe (HRI-EU), Offenbach am Main, Germany (E-mails: thiago.rios@honda-ri.de, stefan.menzel@honda-ri.de, bernhard.sendhoff@honda-ri.de) }

\thanks{Yew-Soon Ong is the Chief Artificial Intelligence Scientist with the Agency for Science, Technology and Research (A*STAR), Singapore, and is also with the Data Science and Artificial Intelligence Research Centre, School of Computer Science and Engineering, Nanyang Technological University (NTU), Singapore (E-mail: ong\_yew\_soon@a-star.edu.sg).}

}
\markboth{Journal of \LaTeX\ Class Files,~Vol.~14, No.~8, August~2015}%
{Shell \MakeLowercase{\textit{et al.}}: Bare Demo of IEEEtran.cls for IEEE Journals}
\maketitle

\begin{abstract}
 
Generative AI relies on statistical learning and local gradient-based guidance, often resulting in artifacts that may be more emulative than genuinely creative. This position paper advances a perspective on Evolutionary Computation (EC) as a paradigm of Natural Generative AI (NGAI), in which stochastic variation, diversification, and natural selection present an alternative foundation for generative exploration and creative synthesis. EC provides an array of variation operators with varied generative capabilities: parent-centric operators induce population distributions analogous to conventional generative models, while disruptive operators can produce structured evolutionary leaps to out-of-distribution artifacts (e.g., programs, designs, or AI agents themselves). This lens reveals a unique theoretical association between certain disruptive variations in multitask EC and a product-of-distributions population structure. A mathematical result derived herein shows that a product structure also occurs under a local second-order approximation of an established generative adversarial program for creative extrapolation, inspired by Martindale's cognitive theory of artistic creation. The simultaneous existence of multiple fitness landscapes in multitask EC inherently moderates selection pressure, allowing these novel artifacts to survive and seed further innovation. \textcolor{black}{Structured disruption and selection moderation are thus argued to be key elements to support creative synthesis. Agnostic to any specific algorithmic realization, the perspective positions NGAI as an emerging engine for open-ended scientific exploration.}
\end{abstract}

\IEEEpeerreviewmaketitle

\section{Introduction}

Generative AI (GenAI) systems use gradient-based learning to model data distributions and synthesize high-quality outputs. Techniques such as Variational Autoencoders (VAEs)~\cite{kingma2013auto,havtorn2021hierarchical,higgins2017beta}, Generative Adversarial Networks (GANs)~\cite{goodfellow2014generative,arjovsky2017wasserstein,shi2023utsgan}, Diffusion Models~\cite{ho2020denoising,song2020score,nichol2021improved} and Large Language Models (LLMs)~\cite{li2023blip,achiam2023gpt,kristiadi2024sober} have been widely applied to various synthesis~\cite{xie2023boxdiff, van2024llamea}, generation~\cite{alayrac2022flamingo, maze2023diffusion} and scientific modeling problems~\cite{Sanchez-Lengeling2018InverseDesign,Mirhoseini2021CircuitDesignAI, miret2025enabling}. However, these methods are arguably more emulative than creative \cite{elgammal2017can}. Their generative capability is largely tied to the distribution of the learned statistical model, with local gradient guidance signals producing outputs that are often biased towards derivatives and interpolations of observed data~\cite{dhariwal2021diffusion, miikkulainen2025neuroevolution}. 


\rev{Innovation in science and engineering critically depends on the ability to explore the extremes of, or extrapolate beyond, known data~\cite{liu2021towards}}. As such, the proclivity of \rev{statistical} GenAI to produce outputs within the bounds of training distributions \cite{dhariwal2021diffusion} limits access to the larger space of possible solutions. 
This issue is particularly acute in domains such as drug discovery \cite{pallikkavaliyaveetil2025small}, materials science~\cite{cheng2025ai}, physical AI and robotics~\cite{wong2026evolutionary}, or complex system design \cite{min2017knowledge} and \rev{optimization~\cite{rabeau2007collaborative}} where prior data may be scarce, costly, or possibly proprietary. The reliance on differentiable functions and gradient-based updates further curtails applicability to designs with black-box, discontinuous, or combinatorial features.

In contrast to conventional GenAI, generative processes in nature---driven by stochastic variation and shaped by Darwinian natural selection---are far more exploratory~\cite{lehman2020surprising}. Random genetic variations foster diversity, leading to the emergence of novel biological forms.  We postulate that this ability of biological evolution to generate novel artifacts is replicable in \emph{in silico} Evolutionary Computation (EC), making it an important computational tool for scientific innovation. Indeed, as highlighted in the Nobel Turing Challenge \cite{kitano2021nobel}, the (often accidental) extension of search spaces beyond their normal scope, coupled with extensive search and optimization, plays a crucial role in the process of discovery. EC naturally adopts such a search-driven generative workflow. As it operates independently of gradient information, evolution excels in black-box or non-differentiable settings. The robustness of evolutionary algorithms in multi-objective and noisy environments further strengthens their practical utility. This presents EC as an alternative and complementary framework for generative exploration and creativity~\cite{miikkulainen2021creative, tian2022modern}. For example, \emph{FunSearch} combines pretrained LLMs with evolutionary search to discover programs for open algorithmic problems \cite{romera2024mathematical}. Its successor, \emph{AlphaEvolve}, orchestrates a pipeline of state-of-the-art LLMs, evolving large pieces of code that may lead to newer scientific discoveries \cite{novikov2025alphaevolve}. Recent work on \emph{Darwin G\"{o}del machines} even demonstrates the prospect of autonomous self-evolution generating a growing archive of AI agents capable of open-ended scientific progress \cite{zhang2026darwin}.

Theoretical advances in information-geometric optimization have revealed a unifying probabilistic foundation for EC, reformulating evolutionary search as a continuous-time optimization process~\cite{ollivier2017information}. This formulation has enabled a deeper formal bridge between EC and generative models. In particular, Estimation of Distribution Algorithms (EDAs) leverage GenAI techniques \rev{for iterative population distribution modeling~\cite{larranaga2001estimation,pelikan2002survey}}. These algorithms continuously refine probabilistic models of the evolving population, gradually increasing the probability of producing higher-quality solutions~\cite{zhang2004convergence}. However, owing to GenAI’s within-distribution bias, generated offspring are largely confined to regions near parental distributions. While this supports stable exploitative search, the population seldom exhibits major leaps or the disruptive generation of out-of-distribution solutions with high functional utility. Realizing the full creative potential of EC thus requires new mechanisms that actively push the boundaries of prior data, thereby enabling faster exploration in open-ended problem solving \cite{miikkulainen2021biological}.

This paper positions EC as a paradigm of \emph{Natural Generative AI} (NGAI), extending the generative capacity of statistical models through stochastic variation, exploratory diversification, and natural selection. 
Unlike the gradient-based guidance of generative models \cite{dhariwal2021diffusion, maze2023diffusion}, evolutionary techniques have been shown to achieve mathematically similar outcomes in a gradient-free manner \cite{wei2025evolvable}. 
Evolutionary algorithms often utilize biologically-inspired genetic operators, iteratively refining populations through recombination and mutation schemes that do not require extensive training on prior data. A wide array of such variation operators has been developed over decades of research, each with its own distinctive generative capabilities. We show that parent-centric variations support localized exploration \rev{\cite{garcia2008global}}, confining offspring generation to known distributions in a manner consistent with conventional GenAI. Disruptive genetic operators enable out-of-distribution leaps, facilitating the discovery of creative solutions outside the boundaries of prior data. While different operators shape the nature of variation, selection pressure remains central in directing the generative dynamics, governing whether the process favors broad exploration or rapid convergence. 

Yet in practice, in silico evolution tends to diverge from biological evolution due to the use of small populations and \emph{strong selection} \cite{miikkulainen2021biological}. Given a narrow focus on solving particular tasks, apparently harmful variations can be prematurely eliminated, leaving large regions of the search space unexplored. This challenge has been recognized in the literature on novelty search \cite{lehman2011novelty} and quality-diversity \cite{pugh2016quality, xiong2026quality}, where explicit behavioral rewards are defined to prevent premature eliminations. \textcolor{black}{We regard such evolutionary approaches as instances of NGAI, with demonstrated ability to generate a diverse repertoire of evolved agent behaviors and designs \cite{cully2015robots, mouret2020quality}.} The idea of Multitask EC (MTEC) offers another unique pathway within NGAI, one that plausibly approximates the dynamics of natural evolution more closely. MTEC, introduced as a model of biocultural multifactorial inheritance \cite{gupta2015multifactorial,ong2016evolutionary}, provides a means of balancing disruptive variation with \emph{moderated selection}. By incorporating multiple fitness landscapes across distinct tasks, \rev{multitasking} reduces the chance of losing out on novel variations---as there is a higher probability of them being useful for at least one task. This consequence has been formalized previously by generalizing Holland's schema theory \cite{holland1992adaptation} to the context of MTEC \cite{gupta2019back}. 

Selection moderation helps preserve greater population diversity, allowing features evolved in one task to transfer across to others \cite{rios2021multitask}. In Section \ref{theory_section}, we present a mathematical result showing that a product-of-distributions structure associated with
certain disruptive transfer mechanisms in MTEC is compatible with a landmark generative adversarial formulation for creative extrapolation \cite{elgammal2017can}. This formulation is inspired by Martindale's cognitive theory of artistic creation, which rewards novelty through controlled deviation from prior patterns \cite{martindale1990clockwork}. Ultimately, our derivation indicates that a training-free recombination rule in MTEC can generate samples consistent with a model of artistic novelty, thus expanding our view of multitask search from an optimization accelerator \cite{lengler2025runtime,libreaking} to a computational engine for broader exploration.

The present paper argues that enabling controlled evolutionary leaps coupled with the preservation of novel generated artifacts is key to realizing the full generative capacity of NGAI. Our main positional claims include:

\begin{itemize}[itemsep=3pt]
\item \textit{Framing EC as a paradigm of NGAI}: EC is interpreted as a stochastic generative process not restricted to prior training datasets. While conventional GenAI stays within trained distributions, NGAI can navigate distributional tails and beyond in a controlled manner.

\item \textit{A mathematical association between creative synthesis and MTEC}: The potential for cross-task feature fusion in MTEC is analyzed through the lens of a product of task-specific distributions. The analysis points to a connection with a generative program inspired by Martindale's cognitive theory of artistic creation. This suggests an alternative way of conceptualizing MTEC, bridging the fields of EC, generative AI and creativity.

\item \textit{The proposal of structured disruption and moderated selection as central principles for creativity}: An illustrative example from the literature is used to showcase the coupling of these effects in a multitask generative design environment. The preservation and structured recombination of solution building blocks across tasks can be seen to produce hybrid designs that depart significantly from the normal scope of either task.
\end{itemize}

The paper is organized as follows. Section II provides preliminaries relating EC and GenAI. In Section III, we make the case that parent-centric operators in EC tend to behave as training-free analogues of conventional generative algorithms, with their application to MTEC inducing a mixture of GenAI models. In Section \ref{sec:ec_disruptive}, we illustrate how the use of disruptive operators can expand the scope of MTEC from mixture models to product distributions. The mathematical link between this product structure and a formal model of computational creativity is then drawn.

\section{GenAI in Evolutionary Computation}

In this section, we present a brief overview of Estimation of Distribution Algorithms (EDAs), a class of evolutionary methods that integrate GenAI with EC in a principled manner. Rather than direct genetic variation of raw candidate solutions in a population, EDAs maintain a parameterized probability distribution over solutions that evolves with time. Each update shifts sampling probability toward regions of the search space $\mathcal{X}$ that contain solutions of higher fitness {$f(\mathbf{x})$}, which can be expressed as the optimization problem: 
 \begin{equation} \arg \max_{\theta} J(\theta) =  \int_{\mathcal{X}} f(\mathbf{x}) \cdot p_{\theta}(\mathbf{x}) \cdot d\mathbf{x}, \quad \text{s.t. } \operatorname{supp}(p_\theta(\mathbf{x})) \subseteq \mathcal{X}. \label{eq:probablistic_ec} \end{equation}
Here, $p_{\theta}(\mathbf{x})$ represents the \rev{probability} distribution underlying the evolving solution population, and $J(\theta)$ is the expected fitness to be maximized \rev{\cite{ollivier2017information}}.

\subsection{Mechanics of Estimation of Distribution Algorithms}

EDAs evolve the population from one generation to the next by building a generative model $p_{\theta}(\mathbf{x})$ of the current parent population and using that to sample offspring. In principle, there is no restriction on the type of generative model used (see Section \ref{sec:genai_eda}), which is why we hereafter refer to these algorithms more broadly as GenAI-EDAs.

\begin{figure}[t]
    \centering    
    \includegraphics[width=0.8\linewidth]{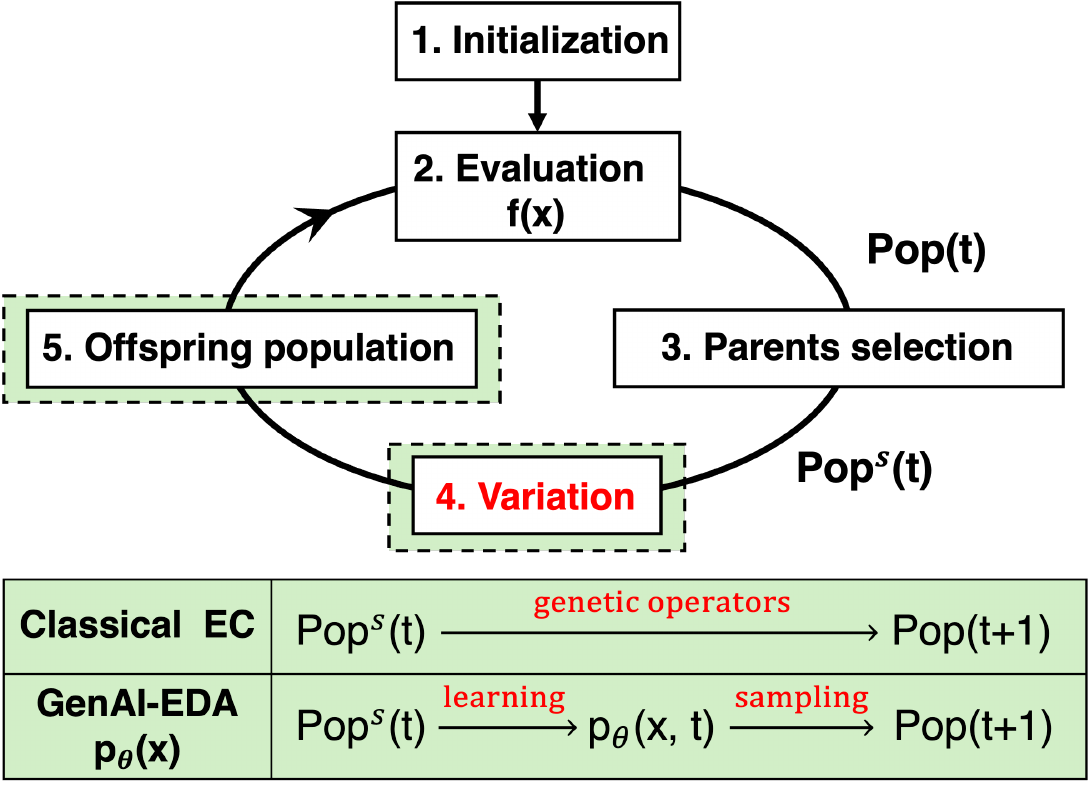}
    \caption{{Illustration of the EDA \rev{workflow where modern GenAI algorithms, as opposed to the biologically-inspired genetic operators of classical EC, are used for solution variation}.}}
    \label{fig:EC_fig3}
    \vspace{-4mm}
\end{figure}

The distinction between EDAs and classical EC is illustrated in Fig.~\ref{fig:EC_fig3}. Let the population of solutions at generation $t$ be denoted as $Pop(t)$ and let the selected parents (favoring those with higher fitness) be denoted as $Pop^s(t)$. Rather than applying explicit genetic operators, EDAs perform the generational transition ${Pop}^{s}(t) \rightarrow {Pop}(t+1)$ as follows:

\begin{itemize}

    \item \textit{Generative modeling (learning)}: build a probability distribution model of the selected parent population:
    \[
    p_{\theta}(\mathbf{x}, t)\ \leftarrow\ Pop^s(t),
    \]

    \item \textit{Variation (sampling):} generate new candidate solutions from the learned model to form the offspring generation:
    \[
    Pop(t+1) \sim p_{\theta}(\mathbf{x}, t).
    \]
    
\end{itemize}

Theoretical results in the literature prove that under certain assumptions (e.g., a large population size), EDAs have the asymptotic convergence property \cite{zhang2004convergence}:
\begin{equation} \lim_{t\rightarrow \infty} \int_{\mathcal{X}} f(\mathbf{x}) \cdot p_{\theta}(\mathbf{x},t) \cdot d\mathbf{x} = f(\mathbf{x}^{*}), \nonumber  
\end{equation} 
where $\mathbf{x}^{*}$ is a global optimum. The basic idea is that as the distribution $p_{\theta}(\mathbf{x})$ evolves, the population mass concentrates near $\mathbf{x}^{*}$. That is, $p_{\theta}(\mathbf{x}, t) \rightarrow \delta(\mathbf{x}-\mathbf{x}^{*})$, where $\delta(\mathbf{x}-\mathbf{x}^{*})$ is the Dirac delta function centered at the optimum.

\subsection{Examples of GenAI-EDAs}\label{sec:genai_eda}
A representative example of this methodological class is the DAE-EDA~\cite{probst2020harmless}, which utilizes a Denoising Autoencoder (DAE) as the generative model \cite{bengio2013generalized} for guided exploration of combinatorial search spaces. Unlike standard EDAs that may experience diversity loss under strong selection, the DAE module introduces controlled randomness during variation, enhancing population diversity and reducing the risk of premature convergence.
\rev{Leveraging techniques such as VAEs \cite{bhattacharjee2019estimation}  or GANs \cite{yu2022enhancing,he2020evolutionary,
dang2024generative}  within the framework of EDAs} gives rise to other algorithms that effectively \rev{guide} localized search while maintaining adaptive exploration.

Evolutionary algorithms have also been studied in conjunction with diffusion models. Zhang et al.~\cite{zhang2025diffusion} view \red{the inverse of the} diffusion process as an evolutionary trajectory (or equivalently, the reverse of evolutionary search as forward diffusion) where a Gaussian distribution iteratively evolves into the target distribution. A related idea has been explored in the context of evolutionary multi-objective optimization as well \cite{yan2024emodm}. While the diffusion evolution algorithm in~\cite{zhang2025diffusion} is model-free (i.e., does not deploy any deep learning architecture), \cite{hartl2026heuristically} incorporated probabilistic diffusion models based on explicit artificial neural networks for generational reproduction in evolutionary algorithms. Interestingly, replacing the diffusion model with a multivariate Gaussian recovers the popular CMA-ES \cite{hansen2001completely}.

\vspace{0.5em}
\noindent\emph{\textbf{Remark 1:} 
GenAI-EDAs integrate probabilistic model-based variations with fitness-guided selection. This approach gradually expands the search scope past parent populations, with the rate of progress largely constrained by the representational limits of GenAI itself. As a result, GenAI-EDAs seldom lead to major evolutionary leaps, instead favoring incremental updates around parent solutions. Although selection pressure pushes the population into unexplored regions of the search space over time, it often takes several generations for distantly located optima to be identified. Consequently, the exploratory potential of EC may remain underutilized when addressing complex optimization or open-ended discovery problems.}


\section{Classical EC as NGAI}\label{sec:ec_as_natgenai}
\textcolor{black}{In this section, we first show that parent-centric genetic operators produce GenAI-like offspring distributions in terms of the regions of the search space that they occupy, leading to similar patterns of incremental exploration around parent populations.} Thereafter, we illustrate that \textcolor{black}{in MTEC, parent-centric recombinations and inter-task solution transfers produce offspring resembling mixtures of generative models.} The potential for creative NGAI, by extending MTEC from mixtures to product distributions, shall be examined in Section~\ref{sec:ec_disruptive}.

\subsection{Parent-centric Operators 
Approximate GenAI-EDAs}\label{sec:ec_parent_centric}
Parent-centric recombination and mutation operators bias offspring to be generated close to their parents~\cite{garcia2008global,ul2020novel}. This behavior is consistent with the previously discussed GenAI-EDA framework, where new solution candidates are sampled from a learned model of the parent distribution and refined through fitness-guided selection.

A representative example of \rev{a} parent-centric operator is the Simulated Binary Crossover (SBX) \cite{Deb1995RealcodedGA, deb2007self}. The extent of stochastic variation in SBX is controlled by a distribution parameter \( \eta \), which balances exploration and exploitation. Bali et al.~\cite{bali2019multifactorial} discussed the generative behavior of such operators under the assumption that evolving populations follow a multivariate Gaussian distribution. To strengthen this presumed connection between parent-centric variations in classical EC and GenAI-EDAs, we conduct a simple empirical study of the two generative processes using more complex \rev{solution distributions. Our study is limited to a two-dimensional space for ease of visualization.} 

\begin{figure*}[t]
    \centering    
    \includegraphics[width=0.95\linewidth]{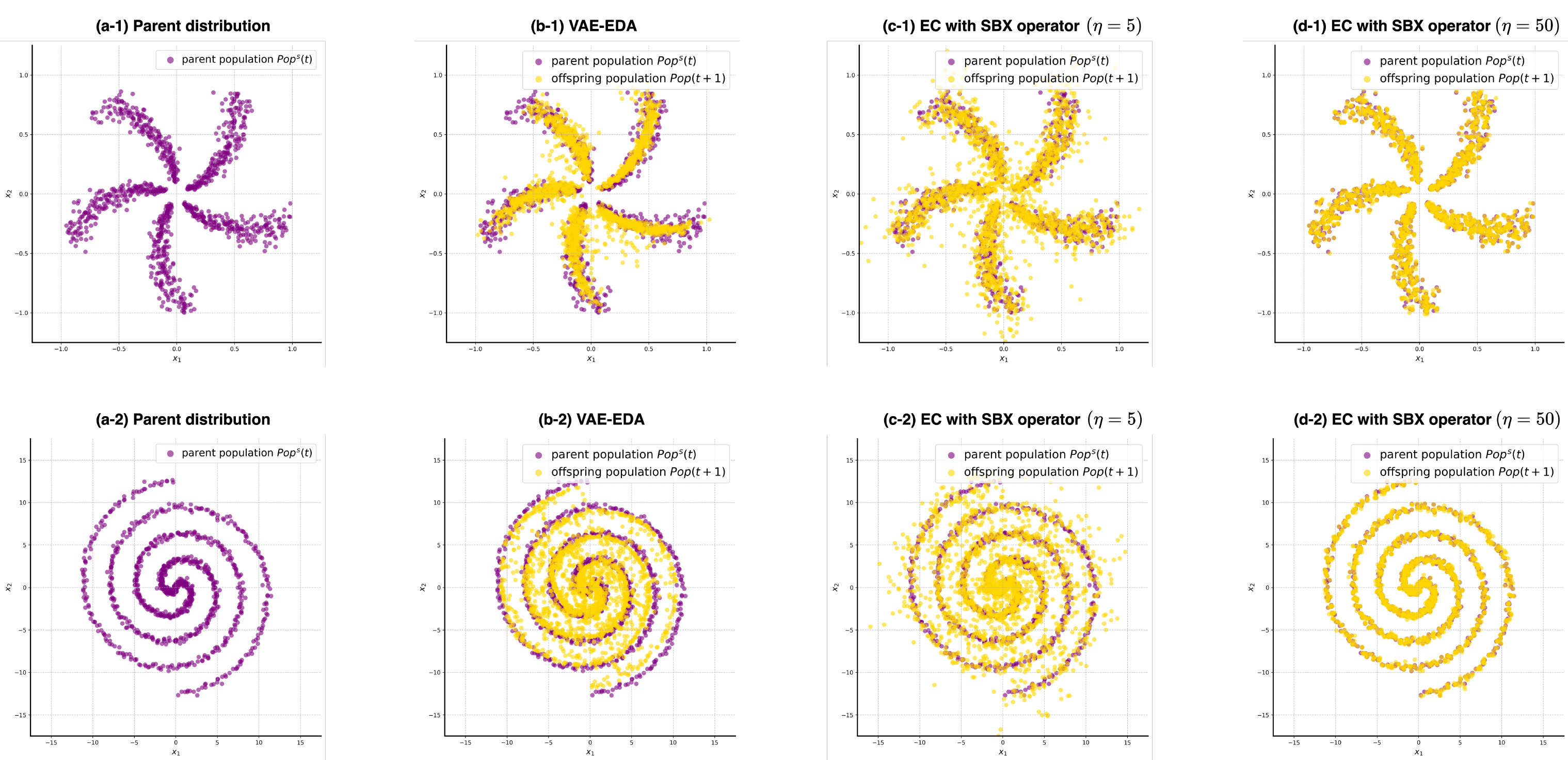}
    \caption{
    \rev{Visualization of offspring generated by GenAI-EDA (with explicit population modeling using a VAE) and the parent-centric \rev{SBX} operator (under different distribution parameter values)}. \rev{Complex parental distributions are used} to assess generative behavior in a general setting. Notably, the SBX operator is gradient-free and does not require the training of a statistical model.}
    \label{fig:EC_fig4_a}
    \vspace{-4mm}
\end{figure*}

Fig.~\ref{fig:EC_fig4_a} depicts two contrived parent \rev{populations} (see subfigures (a-1) and (a-2)) denoted as $Pop^s(t)$. Although such distributions seldom arise in actual EC optimization runs, they provide a testbed for comparing the generative behaviors of different variation schemes. We visualize the offspring populations $Pop(t+1)$ produced by the following procedures:

\begin{itemize}
    
    \item \textit{VAE-EDA}: A conditional VAE, comprising an encoder and decoder implemented as single-hidden-layer neural networks, is trained to model $Pop^s(t)$. The encoder maps 2D parent coordinates and a cluster label (condition) into the parameters of a 32-dimensional latent Gaussian space. The decoder reconstructs 2D coordinates from a sampled latent vector concatenated with the label. Once trained, offspring are generated by sampling from the standard Gaussian prior and decoding under the target condition (see subfigures (b-1) and (b-2) of Fig.~\ref{fig:EC_fig4_a}).
    
    \item \textit{EC with parent-centric recombination}: Offspring are generated by the SBX operator, without the uniform crossover-like variable-swap step used in standard SBX implementations. The operator is applied to randomly selected parent pairs from $Pop^s(t)$. We examine two configurations: $\eta=5$ for more exploratory sampling (see subfigures (c-1) and (c-2) of Fig.~\ref{fig:EC_fig4_a}), and $\eta=50$ for more exploitative behavior (subfigures (d-1) and (d-2)).
    
\end{itemize}

\textcolor{black}{In Fig.~\ref{fig:EC_fig4_a}, both methods produce offspring distributions that remain visually aligned with the parent populations. Both VAE and SBX primarily induce incremental, intra-distributional generation, and in that sense, the results lend empirical support to our conjecture that parent-centric operators exhibit GenAI-like generative behavior.} Furthermore, the SBX distribution index $\eta$ provides tunable control over the exploration–exploitation trade-off, with a larger value of $\eta$ mapping offspring even more closely to their parents. 

\subsection{MTEC Admits a Mixture of GenAI Models}\label{sec:section5_multitaskEC_parent-centric} Despite having access to population-based implicit parallelism, in silico evolutionary algorithms have historically focused narrowly on solving only one problem at a time. Evolution in nature is rarely driven by \rev{such singular fitness measures.} The diversity of biological lifeforms emerges from a multitude of environments and niches, each with its own fitness landscape. Multitask evolution thus offers a closer approximation to the exploratory dynamics of nature by aggregating multiple \rev{problems~\cite{gupta2022guest}}, each associated with its own fitness function. This setup extends the search \rev{process beyond the scope of any single task,} establishing an environment with multiple selection pressures acting on a unified population. 

\begin{figure*}[t]
    \centering    \includegraphics[width=1.0\linewidth]{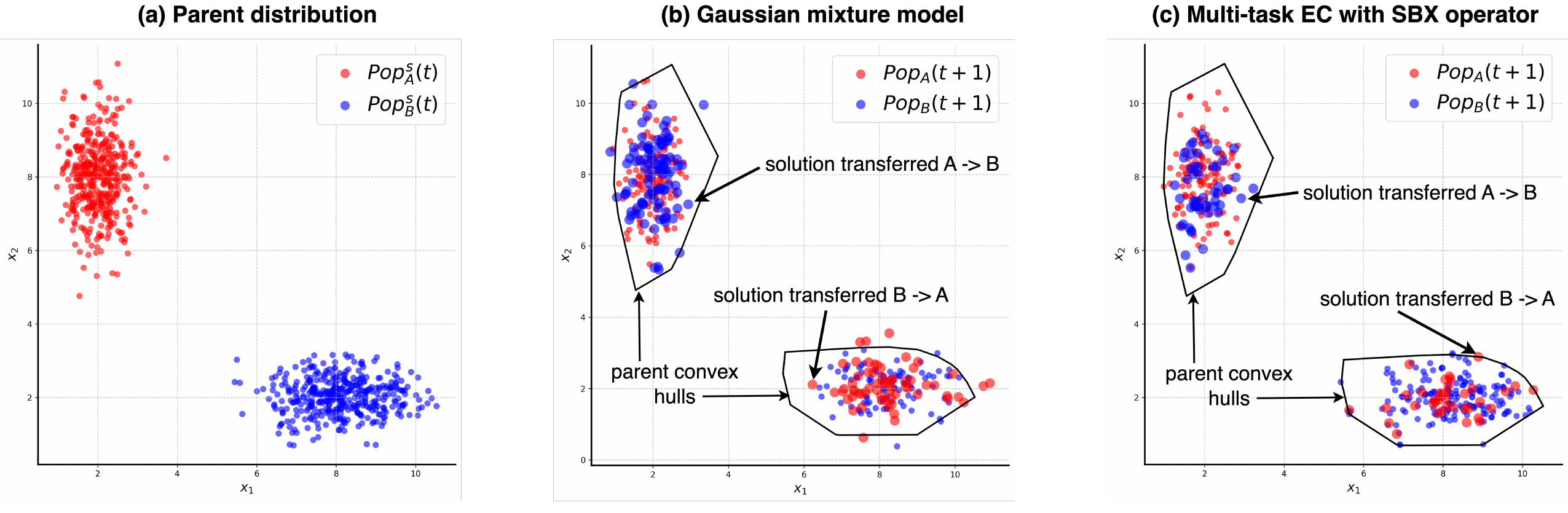} 
    \caption{Visualization of offspring generated by explicit probability  mixture models (Gaussian mixtures) and a parent-centric variation operator (SBX) in a multitask environment. {$Pop^{s}_A(t)$ and $Pop^{s}_B(t)$ represent parent populations for Tasks A and B at generation $t$, while $Pop_A(t+1)$ and $Pop_B(t+1)$ represent the offspring populations for these tasks. The mixing of colors within each convex hull indicates inter-task transfer.}}
    \label{fig:EC_fig5}
    \vspace{-4mm}
\end{figure*}

\textcolor{black}{In MTEC, crossovers may occur between parents drawn from distinct task distributions. An offspring solution resembling a parent from one task could be evaluated on the fitness function of the other, resulting in a form of inter-task solution transfer. Building on our earlier claim that parent-centric operators approximate generative models, we empirically illustrate that parent-centric variations in multitask environments thus lead to offspring that approximate a mixture of task-specific parent distributions.} 

For simplicity, let there be only two tasks with parents drawn from separate models. \textcolor{black}{Note that Gaussian search distributions form the basis of many popular evolutionary algorithms, including the CMA-ES \cite{hansen2001completely}.} As such, solution sets $Pop^s_A(t)$ and $Pop^s_B(t)$,  
\rev{for some hypothetical} Task A and Task B, are independently sampled from anisotropic Gaussians:
\begin{equation} \begin{aligned} p_A(t) \triangleq \mathcal{N}(\boldsymbol{\mu}_A, \boldsymbol{\Sigma}_A), \ p_B(t) \triangleq \mathcal{N}(\boldsymbol{\mu}_B, \boldsymbol{\Sigma}_B), \end{aligned} \label{eq:mixture_components_multitasking} \end{equation}
such that $p_A(t)$ and $p_B(t)$ represent the true underlying distributions of $Pop^s_A(t)$ and $Pop^s_B(t)$, respectively. Without loss of generality, the distributional parameters are defined as:
\begin{equation}
\begin{aligned}
\boldsymbol{\mu}_A &= (2, 8), 
& \boldsymbol{\mu}_B &= (8, 2),
\end{aligned} \nonumber
\label{eq:mixture_components_mean}
\end{equation}
and, 
\begin{equation}
\begin{aligned}
\boldsymbol{\Sigma}_A &= \begin{bmatrix} 0.2 & 0 \\ 0 & 1 \end{bmatrix}, 
& \boldsymbol{\Sigma}_B &= \begin{bmatrix} 1 & 0 \\ 0 & 0.2 \end{bmatrix}.
\end{aligned} \nonumber
\label{eq:mixture_components_covariance}
\end{equation}

\noindent The approaches being compared are configured as follows:
\begin{itemize}
    \item \textit{MTEC with explicit probabilistic modeling}: Bivariate Gaussians $\mathcal{N}(\boldsymbol{\hat{\mu}}_A,\boldsymbol{\hat{\Sigma}}_A)$ and $
    \mathcal{N}(\boldsymbol{\hat{\mu}}_B,\boldsymbol{\hat{\Sigma}}_B)$ are explicitly defined where $\boldsymbol{\hat{\mu}}_A$, $\boldsymbol{\hat{\Sigma}}_A$ and $\boldsymbol{\hat{\mu}}_B$, $\boldsymbol{\hat{\Sigma}}_B$ are the distribution parameters estimated from $Pop^s_A(t)$ and $Pop^s_B(t)$, respectively. Thereafter, two mixture models are constructed with (arbitrary) fixed weights as:
    \begin{equation}\label{eq:mixture_gaussian} \begin{aligned} \hat{p}_A(t+1) \triangleq 0.7 \cdot \mathcal{N}(\boldsymbol{\hat{\mu}}_A, \boldsymbol{\hat{\Sigma}}_A) + 0.3 \cdot \mathcal{N}(\boldsymbol{\hat{\mu}}_B, \boldsymbol{\hat{\Sigma}}_B) \\ \hat{p}_B(t+1) \triangleq 0.5 \cdot \mathcal{N}(\boldsymbol{\hat{\mu}}_A, \boldsymbol{\hat{\Sigma}}_A) + 0.5 \cdot \mathcal{N}(\boldsymbol{\hat{\mu}}_B, \boldsymbol{\hat{\Sigma}}_B). \end{aligned} \end{equation}
    Offspring populations $Pop_A(t+1)$ and $Pop_B(t+1)$ generated by sampling these mixture distributions are depicted in Fig.~\ref{fig:EC_fig5}(b).   
    \item \textit{MTEC with parent-centric recombination}: Offspring are generated using the SBX operator applied to parent pairs sampled from the combined pool $Pop^s_A(t) \cup Pop^s_B(t)$. The generative process follows steps akin to the Multifactorial Evolutionary Algorithm (MFEA) \cite{gupta2015multifactorial, gupta2016multiobjective}: \begin{enumerate*}
    \item \emph{Parent selection}: randomly sample two individuals from the combined parent pool.
    \item \emph{Variation}: apply the SBX operator to generate offspring.
    \item \emph{Task assignment}: if both parents come from the same task, the offspring inherit that task label; else, if parents are from different tasks, the offspring are randomly assigned to either task with equal probability ($p=0.5$). The resulting offspring populations  $Pop_A(t+1)$ and $Pop_B(t+1)$ are visualized in Fig.~\ref{fig:EC_fig5}(c).    
\end{enumerate*}
\end{itemize}

\textcolor{black}{The key takeaway from Fig. \ref{fig:EC_fig5} is that both approaches produce offspring that are largely confined to the convex hulls of the two parent distributions. The mixing of colors within each hull (depicted in Fig.~\ref{fig:EC_fig5}(b) and (c)) highlights the transfer of solutions between tasks; a solution resembling the parents of Task A gets assigned as an offspring of Task B, and vice versa. Although the fraction of such transferred solutions may vary between the two generative processes (owing to arbitrarily chosen mixture weights), the structural consistency between mixture modeling and parent-centric crossovers indicates that they behave analogously in multitask environments. This potential for transfer/mixing is what distinguishes MTEC from its single-task counterpart, reinforcing the former as a generative process capable of intra-task fidelity as well as diversification through inter-task transfer.}

The observed mixing of offspring populations can be used to formalize MTEC. \rev{Considering MTEC with $K$ tasks,} a unified search space $\mathcal{X}$ \rev{is usually defined to encode} candidate solutions for all tasks~\cite{gupta2022half}. Each task is associated with a fitness function, $f_j(\mathbf{x})$, inducing its own selection pressure. Let the population data of the $j$th task be described by the model $p_{\theta_j}(\textbf{x})$. Then, generalizing Eq. (\ref{eq:probablistic_ec}), the expected fitness under a \rev{mixture of generative models can be expressed as:}
\begin{equation} 
   J(\theta_{j}, w_{ij}) = \int_{\mathcal{X}} f_{j}(\psi_{j}^{-1}(\mathbf{x})) \cdot \Bigg[\sum_{i=1}^{K}w_{ij} \cdot p_{\theta_i}(\mathbf{x})\Bigg] \cdot d\mathbf{x}, 
 \label{eq:probablistic_multi_ec} 
\end{equation}
where $\psi_j$ denotes a mapping between the unified space $\mathcal{X}$ and the task-specific \rev{search space \cite{lim2021non}}, and $w_{ij}$ are non-negative mixture coefficients satisfying \textcolor{black}{$\sum_{i=1}^{K} w_{ij} = 1$}. 
Aggregating the expected fitness across all tasks results in the following mathematical formulation for MTEC:
\begin{equation}\label{eq:multitask} 
\centering
\begin{aligned}
&\max_{{\{w_{ij},p_{\theta_j}(\mathbf{x}),\forall{i,j}\}}} \quad \sum_{j=1}^{K}  J(\theta_{j}, w_{ij})
   \\   &\text{s.t. } \operatorname{supp}(p_{\theta_{j}}(\mathbf{x})) \subseteq \mathcal{X}. 
\end{aligned}
\end{equation}
This formulation not only captures the ability of offspring to explore beyond a single parent distribution, but also suggests a mechanism for controlling the extent of inter-task transfers by tuning the mixture weights. When solutions evolved for task \rev{$i$} also perform well at task \rev{$j$}, the aggregated objective favors a larger $w_{ij}$, amplifying inter-task learning. In contrast, if \rev{the} transfer turns out to be detrimental, a smaller $w_{ij}$ is favored, preventing negative interference between tasks \rev{$i$} and \rev{$j$}. 

This connection between multitasking and evolutionary search based on probability mixture models was first drawn in~\cite{bali2019multifactorial}, seeding the adaptive MFEA-II. However, the sampling of a mixture model dictates that the generated offspring are largely tied to the union of task-specific parent hulls. Thus, while MTEC with parent-centric operators broadens exploration outside a single parent distribution, major leaps exploring the extremes of, or extrapolating beyond, the combined parental data are not (yet) enabled.

\section{NGAI for Creative Extrapolation}\label{sec:ec_disruptive}
This section examines how structured disruptions in multitask environments can enable targeted evolutionary leaps beyond the convex hulls of parent distributions. We illustrate empirically that certain disruptive operators—those preferentially recombining higher-frequency genetic material—produce generative behaviors that reasonably approximate a product of task-specific distributions, with mathematical equivalence holding under idealized conditions and prescribed modifications (see Remark 3).

A product of experts facilitates a nonlinear integration of evolved traits, leading to a combinatorial growth in the number of possible task interactions relative to mixture models. Our mathematical derivation shows that a product distribution appears naturally in the solution to a local second-order approximation of an established formulation for creative extrapolation, thereby revealing a latent correspondence between MTEC (with certain disruptive operators) and an approach to computational creativity. Finally, we present an example from the literature on evolutionary generative design that anchors the theory and highlights the role of moderated selection pressure in preserving and refining the generated novelty.

\subsection{Disruption for Structured Evolutionary Leaps}\label{sec:ec_OBSCAN}
Disruptive operators in EC generally introduce substantial variation by breaking usual inheritance patterns, allowing offspring to diverge significantly from parent solutions~\cite{preuss2005counteracting,syswerda1989uniform,eiben1994genetic}. Compared to parent-centric operators, these mechanisms expand exploratory scope and facilitate population diversification outside the boundaries of known data.

A disruptive genetic operator with particularly interesting properties is Occurrence-Based Scanning (OB-Scan)~\cite{eiben1994genetic, wang2010study}. OB-Scan randomly selects a set of parents and, independently for each variable, assigns the offspring corresponding parental values using an occurrence-based rule: more frequent values (i.e., traits with higher population certainty) have higher probability of selection. This majority-informed mechanism promotes a more targeted, variance-reducing, inheritance of features. In the multitask setting, dominant traits across tasks are favorably recombined.

To investigate this behavior, we integrate the OB-Scan operator into the \rev{MFEA-like multitasking} framework. \rev{Refer to Fig.~\ref{fig:EC_fig6}(a), where} parent populations for Tasks A and B are independently sampled from anisotropic Gaussian distributions \( p_A(t) \) and \( p_B(t) \), as defined in Eq. (\ref{eq:mixture_components_multitasking}). Offspring are generated using either the SBX or OB-Scan operator, applied to parent pairs drawn from the combined pool \( Pop^s_A(t) \cup Pop^s_B(t) \), enabling a \rev{balance} between incremental (parent-centric) and disruptive generative dynamics. The offspring generation process follows: \begin{enumerate*} \item \textit{Parent selection}: two individuals are randomly selected from the unified parent pool. \item \textit{Variation}: The SBX or OB-Scan operator is applied with equal probability. {Specifically, for \rev{the} SBX operator, \(\eta\) is set as \(50\). In the OB-Scan operator, as each variable is scanned, the offspring inherits the corresponding parental value with higher probability density, estimated via Gaussian kernel density estimation of the current population distribution (ties are broken randomly).} \item \textit{Task assignment}: offspring directly inherit the task of their parents in \rev{intra-task crossovers}; in inter-task cases, task assignment is randomized as per the MFEA.
\end{enumerate*}

Fig.~\ref{fig:EC_fig6}(b) depicts the generated offspring populations \rev{\( Pop_A(t+1) \) and \( Pop_B(t+1) \).}
This experimental setup yields two key observations. First, it illustrates that inter-task information transfer is preserved, with offspring \rev{being exchanged between the parents' convex hulls}. Second, offspring that lie outside the convex hulls of the respective parent populations also emerge (notice the evolutionary jump marked by a red circle). \emph{Most important, these distant solutions produced by the disruptive OB-Scan operator are not just random, but are biased towards traits of high population certainty by aggregating gene-wise frequency statistics across tasks. Offspring inherit and fuse dominant features originating from both parent clusters (i.e., feature $x_1$ from $Pop_A^s$ and $x_2$ from $Pop_B^s$), thus expanding the generative search space in a structured manner.}
MTEC provides a natural environment for such cross-task feature composition, with each task contributing specialized features evolved for its own fitness landscape.

\subsection{Structured Disruption as a Product Distribution }\label{sec:disruptive_and_product_of_distribution}

\begin{figure*}[t]
    \centering    \includegraphics[width=0.9\linewidth]{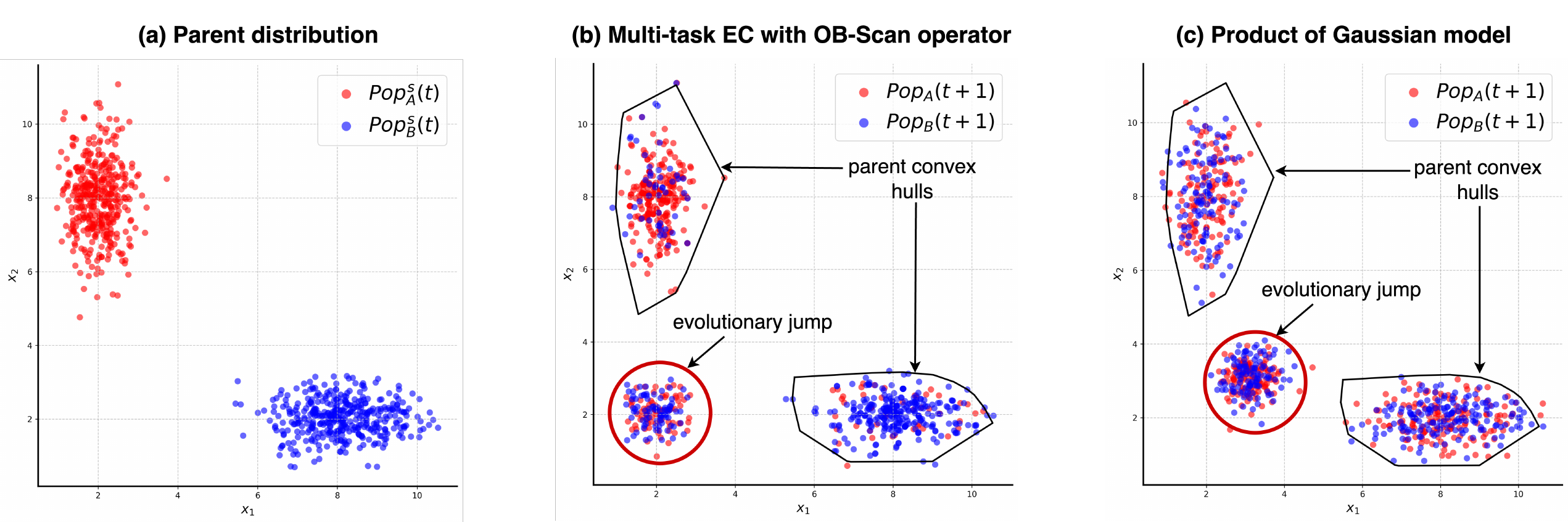}
    \caption{\rev{Visualization of offspring produced in a multitask environment with disruptive genetic operators versus population sampling from a product of generative models. The red circle marks an evolutionary leap outside the parent convex hulls due to the structured disruption of the variation operator.}}
    \label{fig:EC_fig6}
    \vspace{-4mm}
\end{figure*}

Unlike mixture models where data is generated by sampling any one of several underlying experts (analogous to a Boolean \texttt{OR} operation), products of distributions produce data that is jointly consistent with \textit{all} the experts (analogous to an \texttt{AND} operation)~\cite{kant2024identifiability}. Earlier, we presented the OB-Scan operator which biases offspring toward features of high population certainty, thereby targeting the \emph{intersection} of task-specific distributions akin to a product, even though the two are not mathematically equivalent as OB-Scan does not account for inter-variable dependencies. The distinction between mixtures and products is reflected in the difference between the offspring populations in Fig.~\ref{fig:EC_fig5} and Fig.~\ref{fig:EC_fig6}. 
While the offspring in Fig.~\ref{fig:EC_fig5} are attached to the convex hulls of either Task A \texttt{OR} B, the offspring contained inside the red circle in Fig.~\ref{fig:EC_fig6} are formed by a composition of dominant (lower variance) features from both Task A \texttt{AND} B.

In what follows, we illustrate the conceptual association between products and the disruptive OB-Scan operator using our two-dimensional toy example with two axis-aligned multivariate Gaussians (i.e., whose variables are uncorrelated). This association motivates us to think of the product distribution as an abstract approximation of the disruptive genetic operator in the remainder of the paper. In particular, as we will show in Section \ref{theory_section}, the product structure arises naturally in a program for computational creativity.

\vspace{0.5em}
\noindent\emph{\textbf{Remark 2 (Product of Gaussians):}
Let the parent populations of Tasks A and B be modeled as axis-aligned Gaussian distributions $p_A(t) \triangleq \mathcal{N}(\boldsymbol{\mu}_A, \boldsymbol{\Sigma}_A)$ and $p_B(t) \triangleq \mathcal{N}(\boldsymbol{\mu}_B, \boldsymbol{\Sigma}_B)$, respectively, with the distribution parameters defined as:
\begin{equation}
\begin{aligned}
\boldsymbol{\mu}_A &= (\mu_{A1}, \mu_{A2}), 
& \boldsymbol{\mu}_B &= (\mu_{B1}, \mu_{B2}),
\end{aligned} \nonumber
\end{equation}
\begin{equation}
\begin{aligned}
\boldsymbol{\Sigma}_A &= \begin{bmatrix} \sigma_s^2 & 0 \\ 0 & \sigma_l^2 \end{bmatrix}, 
& \boldsymbol{\Sigma}_B &= \begin{bmatrix} \sigma_l^2 & 0 \\ 0 & \sigma_s^2 \end{bmatrix},
\end{aligned} \nonumber
\end{equation}
where $\sigma_s^2 \ll \sigma_l^2$. In other words, $\mu_{A1}$ is the dominant feature of Task A as its population has small variance along this dimension. Similarly, $\mu_{B2}$ is the dominant feature of Task B. Conveniently, the product of Gaussians also happens to be Gaussian \cite{gales2006product}. The offspring population produced by the product is thus described as $\mathcal{N}(\boldsymbol{\mu}_\text{prod}, \boldsymbol{\Sigma}_\text{prod})$ where:
\begin{equation}\label{eq:para_product_gaus}
\boldsymbol{\Sigma}_{\text{prod}} = (\boldsymbol{\Sigma}_A^{-1} + \boldsymbol{\Sigma}_B^{-1})^{-1}, \quad
\boldsymbol{\mu}_{\text{prod}} = \boldsymbol{\Sigma}_{\text{prod}} (\boldsymbol{\Sigma}_A^{-1}\boldsymbol{\mu}_A + \boldsymbol{\Sigma}_B^{-1}\boldsymbol{\mu}_B).
\end{equation}
Plugging the distribution parameters into Eq. (\ref{eq:para_product_gaus}) and setting $\sigma_s^2 / \sigma_l^2 \rightarrow 0$ gives $\boldsymbol{\Sigma}_\text{prod} = \begin{bmatrix} \sigma_s^2 & 0 \\ 0 & \sigma_s^2 \end{bmatrix}$ and $\boldsymbol{\mu}_\text{prod} = (\mu_{A1}, \mu_{B2})$, 
which fuses the dominant traits of both tasks. Each offspring dimension is determined by the parent population with higher certainty (lower variance) in that dimension, approximating a feature-wise \texttt
{AND} operation also associated with the OB-Scan operator.} 

To showcase the OB-Scan and product distribution correspondence, we visually compare offspring populations produced by two generative processes: (1) sampling a weighted sum of a mixture and a product of two task-specific Gaussians, and (2) applying SBX or OB-Scan with equal probability to recombine randomly selected parents (result depicted in Fig.~\ref{fig:EC_fig6}(b)).
Empirical estimates $\mathcal{N}(\boldsymbol{\hat{\mu}}_A, \boldsymbol{\hat{\Sigma}}_A)$ and $\mathcal{N}(\boldsymbol{\hat{\mu}}_B, \boldsymbol{\hat{\Sigma}}_B)$ of the parent distributions are first back-calculated using $Pop^s_A(t)$ and $Pop^s_B(t)$, respectively. $\boldsymbol{\hat{\mu}}_A$, $\boldsymbol{\hat{\Sigma}}_A$ and $\boldsymbol{\hat{\mu}}_B$, $\boldsymbol{\hat{\Sigma}}_B$ are the same as those used in Eq. (\ref{eq:mixture_gaussian}). Offspring for each task are then synthesized by sampling the following composite model:
\begin{equation} 
\begin{aligned} 
\hat{p}_A(t+1) \triangleq w_{11} \cdot \mathcal{N}(\boldsymbol{\hat{\mu}}_A, \boldsymbol{\hat{\Sigma}}_A) + w_{21} \cdot \mathcal{N}(\boldsymbol{\hat{\mu}}_B, \boldsymbol{\hat{\Sigma}}_B) \\+ w_{31} \cdot \mathcal{N}(\boldsymbol{\hat{\mu}}_A, \boldsymbol{\hat{\Sigma}}_A) \times \mathcal{N}(\boldsymbol{\hat{\mu}}_B, \boldsymbol{\hat{\Sigma}}_B), \\ \hat{p}_B(t+1) \triangleq w_{12} \cdot \mathcal{N}(\boldsymbol{\hat{\mu}}_A, \boldsymbol{\hat{\Sigma}}_A) + w_{22} \cdot \mathcal{N}(\boldsymbol{\hat{\mu}}_B, \boldsymbol{\hat{\Sigma}}_B) \\+ w_{32} \cdot \mathcal{N}(\boldsymbol{\hat{\mu}}_A, \boldsymbol{\hat{\Sigma}}_A) \times \mathcal{N}(\boldsymbol{\hat{\mu}}_B, \boldsymbol{\hat{\Sigma}}_B). \end{aligned} 
\end{equation}
We set \(w_{11} = w_{21} = 0.3\), \(w_{12} = w_{22} = 0.3\), and \(w_{31} = w_{32} = 0.4\). The choice of coefficients is arbitrary; \textcolor{black}{alternate choices only alter the fraction of solutions drawn from each component of the mixture without affecting the regions they occupy, and hence do not alter the interpretation of the outcome}.
\(\mathcal{N}(\boldsymbol{\hat{\mu}}_A, \boldsymbol{\hat{\Sigma}}_A) \times \mathcal{N}(\boldsymbol{\hat{\mu}}_B, \boldsymbol{\hat{\Sigma}}_B)\) represents the product whose parameters are calculated based on Eq. (\ref{eq:para_product_gaus}). The sampled data points are depicted in Fig.~\ref{fig:EC_fig6}(c).

The offspring population distributions $\hat{p}_A(t+1)$ and $\hat{p}_B(t+1)$ in Fig.~\ref{fig:EC_fig6}(b) and Fig.~\ref{fig:EC_fig6}(c) reveal qualitative alignment in both structural form and dispersion. 
This observation illustrates our conjecture that effects similar to structured genetic disruption can, at least within the regime of uncorrelated variables, be reasonably approximated by a product of distributions. It facilitates cross-task feature fusion and guides offspring toward regions of higher joint statistical confidence, conceptually parallel to OB-Scan. 

\vspace{0.5em}
\noindent\emph{\textbf{Remark 3 (Modified OB-Scan):}
Notice that the red circle (indicating an evolutionary jump) is slightly shifted in Fig.~\ref{fig:EC_fig6}(c) relative to Fig.~\ref{fig:EC_fig6}(b). This happens because the condition $\sigma_s^2 / \sigma_l^2 \rightarrow 0$ of Remark 2 is an idealization. A modified OB-Scan (modOB-Scan) that emulates the product more closely can be formulated by taking the weighted sum of parental values for each variable---with the weights defined proportionally to the task-specific population certainty (i.e., inverse of the variance)---instead of direct inheritance. Given variable-wise Gaussian densities, the distribution of this sum is mathematically identical to the Gaussian product described by Eq. (\ref{eq:para_product_gaus}).}

Taking the product to be a reasonable approximator of disruptive genetic operators in multitask environments, the probabilistic formulation of MTEC can be expanded by adding product distribution terms into Eq. (\ref{eq:probablistic_multi_ec}). The expected fitness of the $j$th task in this augmented setting becomes:
\begin{equation}
\begin{aligned}
J(\theta_j, w_{ij}, \lambda_n) = \int_{\mathcal{X}} & f_j(\psi_j^{-1}(\mathbf{x})) \cdot \Big[ 
\underbrace{\sum_{i=1}^{K} w_{ij} \cdot p_{\theta_i}(\mathbf{x})}_{\text{(I) Task-wise mixture}} \\
&+ w_{K+1,j} \cdot \underbrace{\sum_{n=1}^{N} \lambda_n 
\cdot \prod_{k \in \mathcal{S}(n)} p_{\theta_k}(\mathbf{x})}_{\text{(II) Product distribution terms}} 
\Big] \, d\mathbf{x},
\end{aligned}
\label{eq:probabilistic_mtec}
\end{equation}
\noindent 
constrained by $\sum_{i=1}^{K} w_{ij} + w_{K+1,j} = 1$. \textcolor{black}{For simplicity of notation, the product terms $\prod_{k \in \mathcal{S}(n)} p_{\theta_k}(\mathbf{x})$ are understood as normalized densities. $\mathcal{S}(n)$ represents a subset of task indices, with $|\mathcal{S}(n)| \geq 2$. This gives rise to $N = 2^{K} - K - 1$  possible higher-order interactions. $\lambda_n \geq 0$ is the weight assigned to each product interaction term and must satisfy $\sum_{n=1}^{N} \lambda_n = 1$.}

\subsection{Linking Structured Disruption and Creative Synthesis}
\label{theory_section}
In a landmark paper in computational creativity \cite{elgammal2017can}, it has been demonstrated that the learning objective of GANs---known for emulating a given distribution---can be modified for creative extrapolation. The new model is labeled a Creative Adversarial Network (CAN). The proposed modification is based on Martindale's cognitive theory of artistic creation \cite{martindale1990clockwork}, which suggests that the arousal potential of art must continually increase, but with the amount of increase bounded from above to avoid any negative reaction (the principle of least effort). While CAN has its origins in generative art, where different artistic styles are blended as a means to increase arousal potential, \emph{the idea is generalized herein by treating the population distribution corresponding to a particular task in MTEC as analogous to the data distribution corresponding to a particular artistic style}. Specifically, we show that solving CAN's mathematical program leads to a generator that can be described in terms of a product of style-/task-specific data distributions, thereby linking to disruptive genetic operators in multitask environments.

As in previous sections, let $p_{\theta_i}(\mathbf{x})$ be the underlying population distribution of the $i$th task, and let $p_d(\mathbf{x}) = \frac{1}{K}\sum_{i=1}^{K}p_{\theta_i}(\mathbf{x})$ denote the distribution of the combined population assuming uniform task representation \cite{bali2019multifactorial}. Each sample $\mathbf{x}$ in the combined population is assumed to be labeled by the task $\tau$ to which it belongs, where $\tau \in \{1, 2, \dots, K\}$. The core objective of CAN can then be written as the following min-max bilevel program \cite{elgammal2017can}:
\begin{equation}\label{CAN_formulation}
\begin{aligned}
\min_{G} \quad
& \mathbb{E}_{\mathbf{x} \sim p_g}
\left[
\log (1 - D_r^*(\mathbf{x}))
-
\textcolor{black}{\gamma}\sum_{\tau=1}^{K}
\frac{1}{K}
\log D_c^*(\tau \mid \mathbf{x})
\right] \\
& \quad
+ \mathbb{E}_{\mathbf{x} \sim p_{d}}
\left[
\log D_r^*(\mathbf{x})
\right] \\
\text{s.t.}\quad
& (D_r^*, D_c^*)
=
\arg\max_{D_r,D_c}
V(G,D_r,D_c).
\end{aligned}
\end{equation}
where $p_g(\mathbf{x})$ denotes the distribution induced by the generator $G$, a style-classification head of the discriminator $D$ estimates the probability $D_c(\tau \mid \mathbf{x})$ that the input $\mathbf{x}$ belongs to task $\tau$, and $D_r(\mathbf{x})$ gives the probability that the input is real (i.e., belongs to $p_d$); \textcolor{black}{$\gamma$ is a parameter that weights the cross-entropy of the style-classification to the uniform distribution.} The CAN discriminator thus has two heads, $D_c$ and $D_r$, optimized by the lower level objective function:
\begin{equation}
\begin{aligned}
V(G, D_r, D_c) = {} & \mathbb{E}_{\mathbf{x}, \tau \sim p_{d}} \left[ \log D_r(\mathbf{x}) + \log D_c(\tau \mid \mathbf{x}) \right] \\
& + \mathbb{E}_{\mathbf{x} \sim p_g} \left[ \log (1 - D_r(\mathbf{x})) \right].
\end{aligned}
\end{equation}

The basic idea is that the discriminator is trained not only to distinguish real samples (which belong to $p_d$) from fake ones (generated by $G$), but also to identify their style-class (i.e., source task). In turn, the generator is trained to produce solution samples that cause the discriminator to make a mistake. In particular, the minimization of the cross-entropy term encourages ambiguity with respect to style classes and is interpreted in CAN as increasing arousal potential, whereas the preservation of realism conforms with the principle of least effort. \textcolor{black}{While $\gamma$ was set to 1 in the original CAN formulation, tuning it allows one to control the extent to which generated samples are encouraged to depart from the true data distribution $p_d$.} Taken together, these two components of the generator's objective serve as a computational implementation of Martindale's theory.

\textcolor{black}{The full mathematical derivation of the generator's optimal distribution is provided in the Appendix. Our key finding is that, under a local second-order approximation, there exist values of $\gamma$ (specifically $\gamma = 1/2$) for which the optimal distribution takes the precise form: }
\begin{equation}
\begin{aligned}
\label{central_result}
p_g^*(\mathbf{x}) \propto  \left( \prod_{\tau=1}^{K} p_{\theta_\tau}(\mathbf{x}) \right)^{1/K},
\end{aligned}
\end{equation}
\textcolor{black}{naturally giving rise to the product of task-specific population distributions. Together with the observation that the disruptive OB-Scan operator relates to product distributions, this indicates that an evolutionary recombination rule requiring no extensive training could, in multitask environments, generate samples consistent with a model of artistic novelty. This result points to a different way of conceptualizing MTEC. Traditional multitask algorithms regard transfer as a joint exploration of \emph{shared} genetic material between tasks. By contrast, viewing MTEC through the lens of probabilistic modeling and CANs suggests that creative divergence could emerge as an intrinsic quality of multitasking through distribution intersection. As search progresses, dominant traits evolved from distinct fitness landscapes can fuse, forming innovative artifacts that may otherwise take longer to discover through single-task generative processes. In what follows, we present an example of such creative extrapolation in generative design.}

\begin{figure*}
    \centering
    \includegraphics[width=0.9 \linewidth]{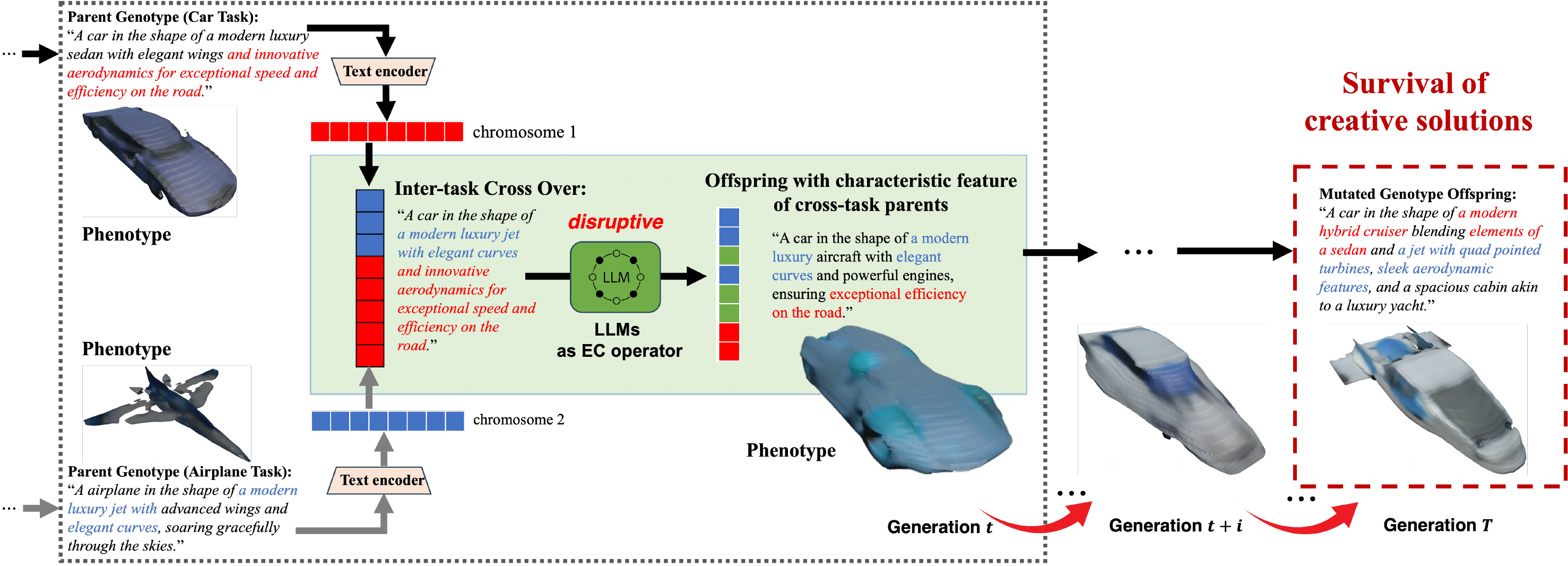}
    \caption{{
        {LLM2TEA for multitasking with LLM-guided natural language solution encoding \cite{wong2025llm2tea}. Generated offspring illustrate gradual out-of-distribution design synthesis through the composition of task-specific features (of cars and airplanes) encoded in the form of textual prompts.}
    }}
    \label{fig:multitasking_crossover}
    \vspace{-4mm}
\end{figure*}

\subsection{\rev{Functional Creativity Through Selection Moderation: An Illustrative Exemplar}}
Creative outputs are not always immediately useful, as they may lack direct functionality for a given task. The strong selection of single-task EC can therefore lead to the premature elimination of novel variations. In contrast, since MTEC brings together multiple fitness landscapes across different tasks, it induces a moderated selection pressure that reduces the chance of premature elimination; there is a higher probability for a randomly generated or evolved solution to be useful for at least one task. This moderation was formally described in \cite{gupta2019back} by generalizing Holland's schema theory \cite{holland1992adaptation} to MTEC, under the assumption of fitness-proportionate selection. Specifically, if $m_j(H,t)$ represents the number of solutions in the population of the $j$th task containing the evolved feature (schema) $H$ at generation $t$, the theory first shows \textcolor{black}{that for short schema}:
\begin{equation}
\begin{aligned}
\mathbb{E}[m_j(H,t+1)] \approx \sum_{i=1}^{K} w_{ij} \cdot m_i(H,t) \cdot \frac{f_i(H,t)}{\bar{f}_i(t)},
\end{aligned}\label{schema_theory}
\end{equation}
where $\mathbb{E}[\cdot]$ is the expectation, $f_i(H,t)$ is the average fitness of solutions containing schema $H$ in the population of the $i$th task with respect to its own fitness landscape, and $\bar{f}_i(t)$ is the average fitness when considering all solutions in that population. Then, denoting the total number of occurrences of schema $H$ in the combined population (union across all tasks) as $m(H,t)$, and assuming the mixture weights to be symmetric (i.e., $w_{ij}=w_{ji}$ as in the MFEA \cite{bali2019multifactorial}), it follows:
\begin{equation}
\begin{aligned}
\mathbb{E}[m(H,t+1)] \approx \sum_{i=1}^{K}  m_i(H,t) \cdot \frac{f_i(H,t)}{\bar{f}_i(t)}.
\end{aligned}\label{schema_theory_final}
\end{equation}
The main insight here is that the frequency of occurrence of a schema can be expected to grow in the population even if $f_j(H,t) < \bar{f}_j(t)$ for some tasks, as long as this is supplemented by sufficiently above-average fitness or high frequency in some other task(s). A stronger corollary to Eq. (\ref{schema_theory_final}) is that for the frequency of an evolved feature to grow, it suffices for its average fitness to be above the population mean for any one task, as long as it is not below average at the others. By providing alternate avenues for suboptimal solutions to survive, MTEC thus supports the emergence and refinement of solutions that are both novel and practical.

\begin{figure*}[t]
    \centering
    \includegraphics[width=0.9 \linewidth]{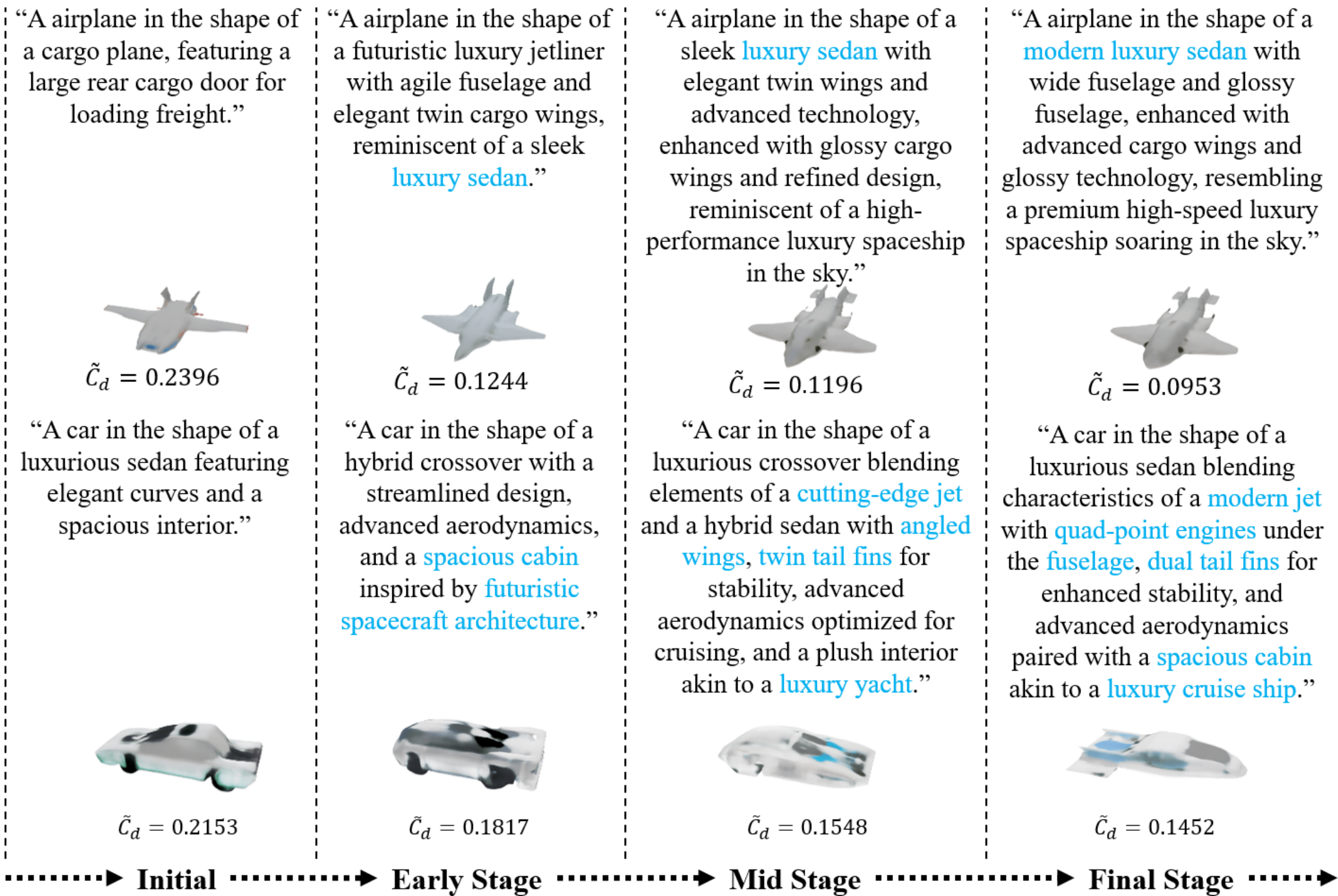}
    \caption{{
        {Trajectory of MTEC during joint aerodynamics optimization of airplanes (top row) and cars (bottom row). $\tilde{C}_d$ represents the drag coefficient of generated designs as the aerodynamic performance indicator to be minimized. As evolution progresses (from left to right), the genotypic traits emerging in the population of one task are transferred to offspring of the other task (highlighted in black) due to the compositional effect of inter-task crossovers. The schema expressed in natural language for blending a sedan with an airplane appears in the early stages of the airplane domain and persists across subsequent stages of both tasks, eventually producing a car with airplane-like features at the bottom right.}
    }}
    \label{fig:multitasking_evolution}
    \vspace{-4mm}
\end{figure*}

To anchor the information-theoretic analysis in Section \ref{theory_section} and the effect of selection pressure moderation, we carry out a post-hoc assessment of an example from the work by Wong et al. \cite{wong2025llm2tea} on LLM-driven Multitask Evolutionary Algorithms (LLM2TEAs). The formulated {\abk{multitasking environment comprises high-dimensional aerodynamic design tasks from two distinct domains: automotive and aerospace}}. Specifically, two design optimization problems were defined, one concerning cars and the other airplanes. The original paper introduced a \emph{Novelty Score} to explicitly quantify the deviation between conventional designs and the unconventional shapes produced by LLM2TEA. Details of this metric are omitted here for brevity, as our focus is on a qualitative assessment of the generated outputs.

LLM2TEA approaches the multitask problem using an integrated agentic AI system, a schematic of which is depicted in Fig.~\ref{fig:multitasking_crossover}. A GPT-family LLM \cite{achiam2023gpt} manages the genetic encoding (in the form of natural language prompts) and executes recombination operations, while a text-to-3D generative model, Shap-E \cite{jun2023shap}, renders the corresponding aerodynamic entities for performance evaluation. The evaluations of aerodynamic drag and lift of the generated designs were carried out
using the open-source computational fluid dynamics software OpenFOAM \cite{weller1998tensorial}. Each task was formulated as a problem that simultaneously aims for high aerodynamic efficiency and strong semantic coherence with domain-specific visual features.
\red{The aggregated objectives (to be minimized) for the car and airplane design tasks are defined respectively as:
\begin{align}
f_{\text{car}}(\textbf{x}) &= \alpha f^{\text{physical}}_{\text{car}}(\textbf{x}) + (1 - \alpha)\cdot(1- f^{\text{visual}}_{\text{car}}(\textbf{x}, \tau_{\text{car}})), \label{eq:car_objective} \\
f_{\text{airplane}}(\textbf{x}) &= \alpha f^{\text{physical}}_{\text{airplane}}(\textbf{x}) + (1 - \alpha) \cdot(1-f^{\text{visual}}_{\text{airplane}}(\textbf{x}, \tau_{\text{airplane}})), \label{eq:airplane_objective}
\end{align}
\noindent where \( f^{\text{physical}}(\cdot) \) denotes the aerodynamic performance, and \( f^{\text{visual}}(\cdot) \) measures semantic alignment with the visual features of the target domain (e.g., "looks like a car" or "looks like an airplane") guided by the visual prompt $\tau$. The predefined hyperparameter \( \alpha \in [0, 1] \) controls the trade-off between physical performance and visual alignment. Full details of the experimental setup are available in \cite{wong2025llm2tea}.}

\red{The free-form expressive capacity of natural language is leveraged to encode design solutions as tokenized textual prompts. For the car and airplane tasks, the genotypes $\mathbf{z}_{\text{car}}$ and $\mathbf{z}_{\text{airplane}}$ follow the templates:
\begin{align}
\mathbf{z}_{\text{car}} &= \text{tokenize}\left(\text{``A $\tau_{\text{car}}$ in the shape of } \langle \boldsymbol{\rho}_{\text{car}} \rangle \text{"}\right), \\
\mathbf{z}_{\text{airplane}} &= \text{tokenize}\left(\text{``An $\tau_{\text{airplane}}$ in the shape of } \langle \boldsymbol{\rho}_{\text{airplane}} \rangle \text{"}\right),
\end{align}
\noindent where $\tau$ defines the task domain and the structural prompt component, $\boldsymbol{\rho}$, is the part of the genotype subject to evolutionary optimization.}

The evolutionary search takes place directly within the space of textual prompts. The fitness of any given prompt $\mathbf{z}$ is determined through a deterministic evaluation pipeline:
\begin{enumerate}
    \item A pre-trained text-to-3D generative model, denoted as $G$, synthesizes a 3D design $\mathbf{x}$ from the input prompt $\mathbf{z}$, i.e., $\mathbf{x} = G(\mathbf{z})$.
    \item The generated design $\mathbf{x}$ is then evaluated using physics simulation software to give $f_{\text{car}}(\mathbf{x})$ or $f_{\text{airplane}}(\mathbf{x})$ depending on the task domain.
\end{enumerate}


A summary of the resulting multitask evolution is presented in Fig.~\ref{fig:multitasking_evolution}. The figure shows the transfer of genotypic traits from one task to the other. For example, the words ``luxurious sedan", initially emerging in the description of cars, eventually cross over to the airplane domain. Likewise, across successive stages of evolution in the car domain, the evolved prompts reveal an incorporation of airplane descriptors. Of relevance to the present discussion is also the survival of the novel designs that emerge as a consequence.

Consider the hybrid design that appears at the bottom-right of Fig.~\ref{fig:multitasking_evolution}, depicting car \texttt{AND} airplane-like features. This design blends a tail fin, a dominant feature of airplanes, onto the body of a car. The fusion is induced by evolved textual prompts that simultaneously contain descriptors of a modern jet and a sedan. Although the underlying mechanics of the LLM's prompt recombination are not interpretable, the observed compositional behavior is reminiscent of the distribution intersection described through Fig. \ref{fig:EC_fig6} under OB-Scan or the product distribution. Interestingly, the schema for blending a sedan with an airplane appears in the early stages of the airplane domain, where its evaluated fitness (measured by drag coefficient $\tilde{C}_d$) is far from optimal. More importantly, the semantic content of that schema can be seen to transfer across tasks and survive through all subsequent stages of both the car and airplane tasks, eventually giving rise to the optimized hybrid. Under the moderation of selection pressure in MTEC, such artifacts showcasing creative disruption have an opportunity to survive and seed further innovation, as illustrated by the refinement of prompts and artifacts from the mid- to final stages of the car domain in particular.

\begin{figure}[t]
    \centering
    \includegraphics[width=0.9 \linewidth]{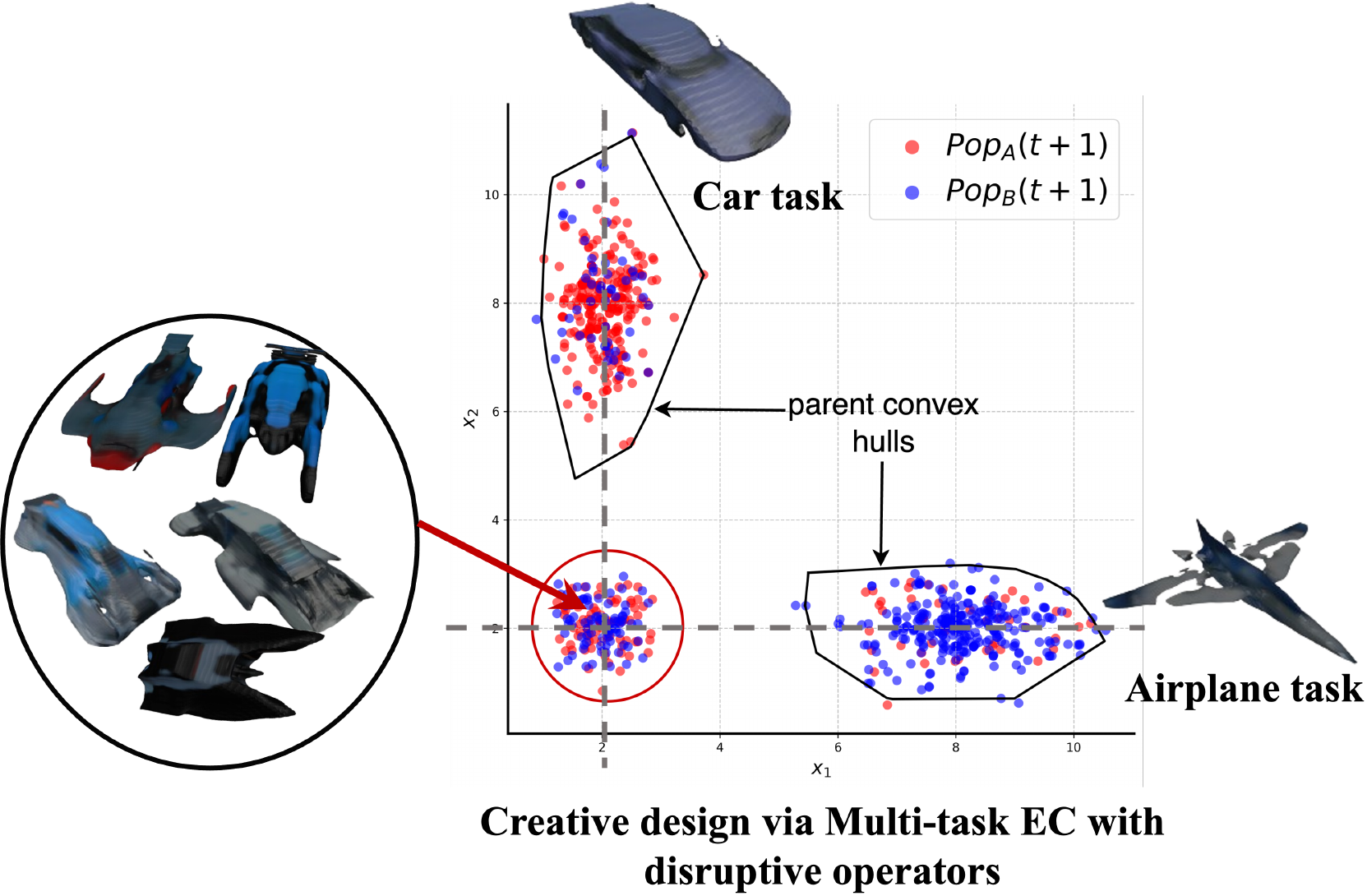}
    \caption{Examples of LLM2TEA-generated designs that intersect distinctive features of car \texttt{AND} airplane shapes, analogous to the evolutionary jump depicted in Fig. \ref{fig:EC_fig6}.}
    \label{fig:more_creative_results}
    \vspace{-4mm}
\end{figure}

A range of other hybrid car-airplane designs were produced by LLM2TEA \cite{wong2025llm2tea}, a subset of which are shown in Fig. \ref{fig:more_creative_results}. These designs lie within the conceptual red circle marking an evolutionary jump, where the dominant features of two distinct parent populations intersect. Note that although in this case these designs already  lay within the support of the text‑to‑3D generator's output distribution, LLM2TEA places a larger probability mass on them by exploring the tails of the distribution. The creativity arises from pushing the boundaries of the distribution (increasing arousal potential) without being jarringly out-of-distribution (Martindale's principle of least effort). The results thus lend support to the notion that LLM-guided semantic crossovers can exhibit generative behavior akin to a product of distributions, with structured disruption in the free-form space of natural language giving rise to outputs that depart strikingly from the design concepts usually represented by the individual tasks.

\section{{Conclusion}}
This paper positions EC as a paradigm of natural generative AI (NGAI), shaped by stochastic variation, creative exploration, and Darwinian selection that accounts for both the fitness and novelty of generated artifacts. \textcolor{black}{The multiplicity and flexibility of generative capabilities offered by EC's varied variation schemes not only enable the discovery of novel scientific solutions and designs (e.g., by Google's AlphaEvolve), but also promise the self-evolution of AI agents that are themselves capable of open-ended scientific progress.} 

\textcolor{black}{Novelty/creativity seeking evolutionary algorithms available in the literature are seen as exemplars of NGAI. In addition to those, this paper particularly argues that EC's multitask extension, MTEC, embodies unique attributes suitable for pushing the boundaries of known data towards creative synthesis.} Key computational elements of MTEC that support our claim include: structured disruption through cross-task feature fusion (synthesizing outputs with high arousal potential), and selection pressure moderation (allowing novel variations to survive and be refined across generations). A recent example of multitasking across a pair of high-dimensional, physics-based generative design tasks illustrates the overall thesis, recasting our view of multitask search from an optimization accelerator to an engine of broader exploration. \textcolor{black}{While our emphasis on creative extrapolation may weaken typical optimization convergence, especially if the optimum happens to lie within known (parental) distributions, it sheds light on \emph{divergence as an intrinsic quality of multitasking} that has largely been overlooked to date.} 

Our analysis suggests that the generative behavior of certain disruptive transfer mechanisms in MTEC has similarities with a product of distributions. A derived mathematical result indicates that this product structure also emerges in the optimal solution to a local second-order approximation of a generative adversarial formulation for creative extrapolation. The formulation describes a Creative Adversarial Network (CAN) inspired by Martindale's cognitive theory of artistic creation \cite{elgammal2017can}, which rewards novelty through controlled deviation from prior patterns. Our derivation thus reveals a mathematical correspondence between multitasking and computational creativity, bridging the fields of EC, generative AI and creativity in a manner not fully acknowledged in the past. The moderation of selection pressure in MTEC is further supported by the generalization of Holland's schema theory to multitasking, lending additional theoretical grounding to our claims.

The work illuminates promising directions for future algorithmic research within the framework of NGAI. While parent-centric genetic operators offer a training-free means of approximating conventional GenAI algorithms, the disruptive OB-Scan in MTEC is associated with a product-of-distributions, as illustrated with axis-aligned distributions. In Remark 3, Section \ref{sec:disruptive_and_product_of_distribution}, a \emph{modOB-Scan} scheme is formulated that is mathematically identical to the product, when limiting to variable-wise Gaussian densities. In general, however, these generative processes are not the same, despite both inducing a compositional effect to fuse dominant traits from distinct tasks (akin to a Boolean \texttt{AND} operation).  For an evolutionary algorithm to fully embody our proposed theory---for example, without having to train a CAN every generation for iterative solution sampling---the development of training-free operators that approach the derived distribution in Eq. (\ref{central_result}) more closely will be of significant value.

\section*{Declaration of conflict of interest}
The authors declare that they have no known competing financial interests or personal relationships that could have appeared to influence the work reported in this paper.

{
\color{black}
\appendix
Here, we present the full derivation of the generator's optimal distribution in the CAN formulation.
From Theorem 1 in \cite{goodfellow2014generative}, it is known that:
\begin{equation}
\begin{aligned}
&\mathbb{E}_{\mathbf{x} \sim p_g}
\left[
\log (1 - D_r^*(\mathbf{x}))
\right]
+
\mathbb{E}_{\mathbf{x} \sim p_{d}}
\left[
\log D_r^*(\mathbf{x})
\right] \\
&\qquad
= -\log(4) + 2\,JSD(p_d \parallel p_g),
\end{aligned}
\end{equation}
where $JSD$ is the Jensen–Shannon divergence between the generator's induced distribution and the data distribution. Further, the prediction of the optimal style-classifier shall match the true posterior, implying:
\begin{equation}
\begin{aligned}
D_c^*(\tau \mid \mathbf{x})
=
\frac{p_{\theta_\tau}(\mathbf{x})}
{\sum_{j=1}^{K} p_{\theta_j}(\mathbf{x})}.
\end{aligned}
\end{equation}

Substituting these results, which solve the discriminator's problem at the lower level, into the generator's objective in Eq. (\ref{CAN_formulation}) gives the following reduced mathematical program:
\begin{equation}
\begin{aligned}
\min_{G} 2JSD(p_d \parallel p_g) - \gamma \;\mathbb{E}_{\mathbf{x} \sim p_g} \left[ \sum_{\tau=1}^{K} \frac{1}{K} \log \frac{p_{\theta_\tau}(\mathbf{x})}{p_d(\mathbf{x})} \right],
\end{aligned}
\end{equation}
where the constant terms are dropped.

The realism condition in CAN, which captures Martindale's principle of least effort, motivates analyzing the generator within a bounded divergence of the data distribution. This encapsulates the idea that creativity should not be jarringly out-of-distribution, as generated outputs that are too radical can elicit unfavorable responses. As such, we focus our analysis to a local neighborhood of $p_d$ where the Jensen–Shannon divergence admits the second-order approximation \cite{lin1991divergence}:
\begin{equation}
\begin{aligned}
JSD(p_d \parallel p_g) \approx \frac{1}{4}KL(p_g \parallel p_d),
\end{aligned}
\end{equation}
where $KL(p_g \parallel p_d) = \int p_g(\mathbf{x}) \log \frac{p_g(\mathbf{x})}{p_d(\mathbf{x})} \, d\mathbf{x}$ is the Kullback-Leibler divergence. Expanding the generator's objective gives:
\begin{equation}\label{localized_CAN_formulation}
\begin{aligned}
\min_{G}\;
&\frac{1}{2}
\int p_g(\mathbf{x})
\log \frac{p_g(\mathbf{x})}{p_d(\mathbf{x})}
\, d\mathbf{x} \\
&-
\int p_g(\mathbf{x})
\left[\gamma
\sum_{\tau=1}^{K}
\frac{1}{K}
\log
\frac{p_{\theta_\tau}(\mathbf{x})}
{p_d(\mathbf{x})}
\right]
\, d\mathbf{x}.
\end{aligned}
\end{equation}
Notice that choosing $\gamma = 1/2$ causes all $p_d$ terms in Eq. (\ref{localized_CAN_formulation}) to cancel. After some algebraic rearrangements, we get the following by grouping of the logarithmic terms under a single integral:
\begin{equation}
\begin{aligned}
\min_{G}\;
&\frac{1}{2}
\int p_g(\mathbf{x})
\Biggl[
\log p_g(\mathbf{x})
-
\log\!\left(
\left(
\prod_{\tau=1}^{K}
p_{\theta_\tau}(\mathbf{x})
\right)^{1/K}
\right)
\Biggr]
\, d\mathbf{x}.
\end{aligned}
\end{equation}
which is equivalent to the minimization of a direct divergence form:
\begin{equation}
\begin{aligned}
\label{KL_divergence}
\min_{G} \frac{1}{2} \int p_g(\mathbf{x}) \log \left(\frac{p_g(\mathbf{x})}{ \left( \prod_{\tau=1}^{K} p_{\theta_\tau}(\mathbf{x}) \right)^{1/K}} \right) d\mathbf{x}.
\end{aligned}
\end{equation}

The integral in Eq. (\ref{KL_divergence}) gives the KL divergence between $p_g(\mathbf{x})$ and $\frac{1}{Z} \left( \prod_{\tau=1}^{K} p_{\theta_\tau}(\mathbf{x}) \right)^{1/K}$, up to an additive constant $-\log(Z)$, where $Z$ is a normalization term to ensure a valid density. From the fact that the minimum value of the KL divergence (equal to zero) occurs when the two distributions coincide \cite{kullback1951information}, it follows that the optimal distribution induced by the generator must take the form:
\begin{equation}
\begin{aligned}
p_g^*(\mathbf{x}) = \frac{1}{Z} \left( \prod_{\tau=1}^{K} p_{\theta_\tau}(\mathbf{x}) \right)^{1/K}.
\end{aligned}
\end{equation}
}

\bibliographystyle{IEEEtran}
\bibliography{IEEE_CIM_NatGenAI}

\begin{thebibliography}{10}
\providecommand{\url}[1]{#1}
\csname url@samestyle\endcsname
\providecommand{\newblock}{\relax}
\providecommand{\bibinfo}[2]{#2}
\providecommand{\BIBentrySTDinterwordspacing}{\spaceskip=0pt\relax}
\providecommand{\BIBentryALTinterwordstretchfactor}{4}
\providecommand{\BIBentryALTinterwordspacing}{\spaceskip=\fontdimen2\font plus
\BIBentryALTinterwordstretchfactor\fontdimen3\font minus \fontdimen4\font\relax}
\providecommand{\BIBforeignlanguage}[2]{{%
\expandafter\ifx\csname l@#1\endcsname\relax
\typeout{** WARNING: IEEEtran.bst: No hyphenation pattern has been}%
\typeout{** loaded for the language `#1'. Using the pattern for}%
\typeout{** the default language instead.}%
\else
\language=\csname l@#1\endcsname
\fi
#2}}
\providecommand{\BIBdecl}{\relax}
\BIBdecl

\bibitem{kingma2013auto}
D.~P. Kingma and M.~Welling, ``Auto-encoding variational bayes,'' \emph{arXiv preprint arXiv:1312.6114}, 2013.

\bibitem{havtorn2021hierarchical}
J.~D. Havtorn, J.~Frellsen, S.~Hauberg, and L.~Maal{\o}e, ``Hierarchical vaes know what they don’t know,'' in \emph{International Conference on Machine Learning}.\hskip 1em plus 0.5em minus 0.4em\relax PMLR, 2021, pp. 4117--4128.

\bibitem{higgins2017beta}
I.~Higgins, L.~Matthey, A.~Pal, C.~Burgess, X.~Glorot, M.~Botvinick, S.~Mohamed, and A.~Lerchner, ``beta-vae: Learning basic visual concepts with a constrained variational framework,'' in \emph{International conference on learning representations}, 2017.

\bibitem{goodfellow2014generative}
I.~Goodfellow, J.~Pouget-Abadie, M.~Mirza, B.~Xu, D.~Warde-Farley, S.~Ozair, A.~Courville, and Y.~Bengio, ``Generative adversarial nets,'' \emph{Advances in neural information processing systems}, vol.~27, 2014.

\bibitem{arjovsky2017wasserstein}
M.~Arjovsky, S.~Chintala, and L.~Bottou, ``Wasserstein generative adversarial networks,'' in \emph{International conference on machine learning}.\hskip 1em plus 0.5em minus 0.4em\relax PMLR, 2017, pp. 214--223.

\bibitem{shi2023utsgan}
Y.~Shi, X.~Zhou, P.~Liu, and I.~W. Tsang, ``Utsgan: Unseen transition suss gan for transition-aware image-to-image translation,'' \emph{arXiv preprint arXiv:2304.11955}, 2023.

\bibitem{ho2020denoising}
J.~Ho, A.~Jain, and P.~Abbeel, ``Denoising diffusion probabilistic models,'' \emph{Advances in neural information processing systems}, vol.~33, pp. 6840--6851, 2020.

\bibitem{song2020score}
Y.~Song, J.~Sohl-Dickstein, D.~P. Kingma, A.~Kumar, S.~Ermon, and B.~Poole, ``Score-based generative modeling through stochastic differential equations,'' \emph{arXiv preprint arXiv:2011.13456}, 2020.

\bibitem{nichol2021improved}
A.~Q. Nichol and P.~Dhariwal, ``Improved denoising diffusion probabilistic models,'' in \emph{International conference on machine learning}.\hskip 1em plus 0.5em minus 0.4em\relax PMLR, 2021, pp. 8162--8171.

\bibitem{li2023blip}
J.~Li, D.~Li, S.~Savarese, and S.~Hoi, ``Blip-2: Bootstrapping language-image pre-training with frozen image encoders and large language models,'' in \emph{International conference on machine learning}.\hskip 1em plus 0.5em minus 0.4em\relax PMLR, 2023, pp. 19\,730--19\,742.

\bibitem{achiam2023gpt}
J.~Achiam, S.~Adler, S.~Agarwal, L.~Ahmad, I.~Akkaya, F.~L. Aleman, D.~Almeida, J.~Altenschmidt, S.~Altman, S.~Anadkat \emph{et~al.}, ``Gpt-4 technical report,'' \emph{arXiv preprint arXiv:2303.08774}, 2023.

\bibitem{kristiadi2024sober}
A.~Kristiadi, F.~Strieth-Kalthoff, M.~Skreta, P.~Poupart, A.~Aspuru-Guzik, and G.~Pleiss, ``A sober look at llms for material discovery: Are they actually good for bayesian optimization over molecules?'' in \emph{International Conference on Machine Learning}.\hskip 1em plus 0.5em minus 0.4em\relax PMLR, 2024, pp. 25\,603--25\,622.

\bibitem{xie2023boxdiff}
J.~Xie, Y.~Li, Y.~Huang, H.~Liu, W.~Zhang, Y.~Zheng, and M.~Z. Shou, ``Boxdiff: Text-to-image synthesis with training-free box-constrained diffusion,'' in \emph{Proceedings of the IEEE/CVF International Conference on Computer Vision}, 2023, pp. 7452--7461.

\bibitem{van2024llamea}
N.~Van~Stein and T.~B{\"a}ck, ``Llamea: A large language model evolutionary algorithm for automatically generating metaheuristics,'' \emph{IEEE Transactions on Evolutionary Computation}, vol.~29, no.~2, pp. 331--345, 2024.

\bibitem{alayrac2022flamingo}
J.-B. Alayrac, J.~Donahue, P.~Luc, A.~Miech, I.~Barr, Y.~Hasson, K.~Lenc, A.~Mensch, K.~Millican, M.~Reynolds \emph{et~al.}, ``Flamingo: a visual language model for few-shot learning,'' \emph{Advances in neural information processing systems}, vol.~35, pp. 23\,716--23\,736, 2022.

\bibitem{maze2023diffusion}
F.~Maz{\'e} and F.~Ahmed, ``Diffusion models beat gans on topology optimization,'' in \emph{Proceedings of the AAAI conference on artificial intelligence}, vol.~37, no.~8, 2023, pp. 9108--9116.

\bibitem{Sanchez-Lengeling2018InverseDesign}
B.~Sánchez-Lengeling and A.~Aspuru-Guzik, ``Inverse molecular design using machine learning: Generative models for matter engineering,'' \emph{Science}, vol. 361, no. 6400, pp. 360--365, 2018.

\bibitem{Mirhoseini2021CircuitDesignAI}
A.~Mirhoseini, A.~Goldie, M.~Yazgan, J.~W. Jiang, E.~Songhori, S.~Wang, Y.-J. Lee, E.~Johnson, O.~Pathak, A.~Nova \emph{et~al.}, ``A graph placement methodology for fast chip design,'' \emph{Nature}, vol. 594, no. 7862, pp. 207--212, 2021.

\bibitem{miret2025enabling}
S.~Miret and N.~A. Krishnan, ``Enabling large language models for real-world materials discovery,'' \emph{Nature Machine Intelligence}, vol.~7, no.~7, pp. 991--998, 2025.

\bibitem{elgammal2017can}
A.~Elgammal, B.~Liu, M.~Elhoseiny, and M.~Mazzone, ``Can: Creative adversarial networks generating “art” by learning about styles and deviating from style norms,'' in \emph{8th International Conference on Computational Creativity, ICCC 2017}.\hskip 1em plus 0.5em minus 0.4em\relax Georgia Institute of Technology, 2017.

\bibitem{dhariwal2021diffusion}
P.~Dhariwal and A.~Nichol, ``Diffusion models beat gans on image synthesis,'' \emph{Advances in neural information processing systems}, vol.~34, pp. 8780--8794, 2021.

\bibitem{miikkulainen2025neuroevolution}
R.~Miikkulainen, ``Neuroevolution insights into biological neural computation,'' \emph{Science}, vol. 387, no. 6735, p. eadp7478, 2025.

\bibitem{liu2021towards}
J.~Liu, Z.~Shen, Y.~He, X.~Zhang, R.~Xu, H.~Yu, and P.~Cui, ``Towards out-of-distribution generalization: A survey,'' \emph{arXiv preprint arXiv:2108.13624}, 2021.

\bibitem{pallikkavaliyaveetil2025small}
N.~Pallikkavaliyaveetil and S.~Chandrasekaran, ``Small data, big challenges: Machine-and deep-learning strategies for data-limited drug discovery,'' \emph{Advanced drug delivery reviews}, p. 115762, 2025.

\bibitem{cheng2025ai}
M.~Cheng, C.-L. Fu, R.~Okabe, A.~Chotrattanapituk, A.~Boonkird, N.~T. Hung, and M.~Li, ``Ai-driven materials design: a mini-review,'' \emph{arXiv preprint arXiv:2502.02905}, 2025.

\bibitem{wong2026evolutionary}
J.~C. Wong, A.~Gupta, C.~C. Ooi, P.-H. Chiu, J.~Liu, and Y.-S. Ong, ``Evolutionary optimization of physics-informed neural networks: Evo-pinn frontiers and opportunities,'' \emph{IEEE Computational Intelligence Magazine}, vol.~21, no.~1, pp. 16--36, 2026.

\bibitem{min2017knowledge}
A.~T.~W. Min, R.~Sagarna, A.~Gupta, Y.-S. Ong, and C.~K. Goh, ``Knowledge transfer through machine learning in aircraft design,'' \emph{IEEE Computational Intelligence Magazine}, vol.~12, no.~4, pp. 48--60, 2017.

\bibitem{rabeau2007collaborative}
S.~Rabeau, P.~D{\'e}pinc{\'e}, and F.~Bennis, ``Collaborative optimization of complex systems: a multidisciplinary approach,'' \emph{International Journal on Interactive Design and Manufacturing (IJIDeM)}, vol.~1, no.~4, pp. 209--218, 2007.

\bibitem{lehman2020surprising}
J.~Lehman, J.~Clune, D.~Misevic, C.~Adami, L.~Altenberg, J.~Beaulieu, P.~J. Bentley, S.~Bernard, G.~Beslon, D.~M. Bryson \emph{et~al.}, ``The surprising creativity of digital evolution: A collection of anecdotes from the evolutionary computation and artificial life research communities,'' \emph{Artificial life}, vol.~26, no.~2, pp. 274--306, 2020.

\bibitem{kitano2021nobel}
H.~Kitano, ``Nobel turing challenge: creating the engine for scientific discovery,'' \emph{NPJ systems biology and applications}, vol.~7, no.~1, p.~29, 2021.

\bibitem{miikkulainen2021creative}
R.~Miikkulainen, ``Creative ai through evolutionary computation: principles and examples,'' \emph{SN Computer Science}, vol.~2, no.~3, p. 163, 2021.

\bibitem{tian2022modern}
Y.~Tian and D.~Ha, ``Modern evolution strategies for creativity: Fitting concrete images and abstract concepts,'' in \emph{International conference on computational intelligence in music, sound, art and design (part of evostar)}.\hskip 1em plus 0.5em minus 0.4em\relax Springer, 2022, pp. 275--291.

\bibitem{romera2024mathematical}
B.~Romera-Paredes, M.~Barekatain, A.~Novikov, M.~Balog, M.~P. Kumar, E.~Dupont, F.~J. Ruiz, J.~S. Ellenberg, P.~Wang, O.~Fawzi \emph{et~al.}, ``Mathematical discoveries from program search with large language models,'' \emph{Nature}, vol. 625, no. 7995, pp. 468--475, 2024.

\bibitem{novikov2025alphaevolve}
A.~Novikov, N.~V{\~u}, M.~Eisenberger, E.~Dupont, P.-S. Huang, A.~Z. Wagner, S.~Shirobokov, B.~Kozlovskii, F.~J. Ruiz, A.~Mehrabian \emph{et~al.}, ``Alphaevolve: A coding agent for scientific and algorithmic discovery,'' \emph{arXiv preprint arXiv:2506.13131}, 2025.

\bibitem{zhang2026darwin}
J.~Zhang, S.~Hu, C.~Lu, R.~Lange, and J.~Clune, ``Darwin g{\"o}del machine: Open-ended evolution of self-improving agents,'' in \emph{International Conference on Learning Representations}, vol. 2026, 2026, pp. 104\,223--104\,294.

\bibitem{ollivier2017information}
Y.~Ollivier, L.~Arnold, A.~Auger, and N.~Hansen, ``Information-geometric optimization algorithms: A unifying picture via invariance principles,'' \emph{Journal of Machine Learning Research}, vol.~18, no.~18, pp. 1--65, 2017.

\bibitem{larranaga2001estimation}
P.~Larra{\~n}aga and J.~A. Lozano, \emph{Estimation of distribution algorithms: A new tool for evolutionary computation}.\hskip 1em plus 0.5em minus 0.4em\relax Springer Science \& Business Media, 2001, vol.~2.

\bibitem{pelikan2002survey}
M.~Pelikan, D.~E. Goldberg, and F.~G. Lobo, ``A survey of optimization by building and using probabilistic models,'' \emph{Computational optimization and applications}, vol.~21, pp. 5--20, 2002.

\bibitem{zhang2004convergence}
Q.~Zhang and H.~Muhlenbein, ``On the convergence of a class of estimation of distribution algorithms,'' \emph{IEEE Transactions on evolutionary computation}, vol.~8, no.~2, pp. 127--136, 2004.

\bibitem{miikkulainen2021biological}
R.~Miikkulainen and S.~Forrest, ``A biological perspective on evolutionary computation,'' \emph{Nature Machine Intelligence}, vol.~3, no.~1, pp. 9--15, 2021.

\bibitem{wei2025evolvable}
Z.~Wei, C.~C. Ooi, A.~Gupta, J.~C. Wong, P.-H. Chiu, S.~X.~W. Toh, and Y.-S. Ong, ``Evolvable conditional diffusion,'' in \emph{Proceedings of the Thirty-Fourth International Joint Conference on Artificial Intelligence}, 2025, pp. 6542--6551.

\bibitem{garcia2008global}
C.~Garc{\'\i}a-Mart{\'\i}nez, M.~Lozano, F.~Herrera, D.~Molina, and A.~M. S{\'a}nchez, ``Global and local real-coded genetic algorithms based on parent-centric crossover operators,'' \emph{European journal of operational research}, vol. 185, no.~3, pp. 1088--1113, 2008.

\bibitem{lehman2011novelty}
J.~Lehman and K.~O. Stanley, ``Novelty search and the problem with objectives,'' in \emph{Genetic programming theory and practice IX}.\hskip 1em plus 0.5em minus 0.4em\relax Springer, 2011, pp. 37--56.

\bibitem{pugh2016quality}
J.~K. Pugh, L.~B. Soros, and K.~O. Stanley, ``Quality diversity: A new frontier for evolutionary computation,'' \emph{Frontiers in Robotics and AI}, vol.~3, p.~40, 2016.

\bibitem{xiong2026quality}
A.~K.~B. Xiong, M.~Wong, J.~Liu, C.~Chen, and Y.-S. Ong, ``Quality-diversity llm for generative design,'' in \emph{AI4X Accelerate Conference}, 2026.

\bibitem{cully2015robots}
A.~Cully, J.~Clune, D.~Tarapore, and J.-B. Mouret, ``Robots that can adapt like animals,'' \emph{Nature}, vol. 521, no. 7553, pp. 503--507, 2015.

\bibitem{mouret2020quality}
J.-B. Mouret and G.~Maguire, ``Quality diversity for multi-task optimization,'' in \emph{Proceedings of the 2020 Genetic and Evolutionary Computation Conference}, 2020, pp. 121--129.

\bibitem{gupta2015multifactorial}
A.~Gupta, Y.-S. Ong, and L.~Feng, ``Multifactorial evolution: Toward evolutionary multitasking,'' \emph{IEEE Transactions on Evolutionary Computation}, vol.~20, no.~3, pp. 343--357, 2015.

\bibitem{ong2016evolutionary}
Y.-S. Ong and A.~Gupta, ``Evolutionary multitasking: A computer science view of cognitive multitasking,'' \emph{Cognitive Computation}, vol.~8, pp. 125--142, 2016.

\bibitem{holland1992adaptation}
J.~H. Holland, \emph{Adaptation in natural and artificial systems: an introductory analysis with applications to biology, control, and artificial intelligence}.\hskip 1em plus 0.5em minus 0.4em\relax MIT press, 1992.

\bibitem{gupta2019back}
A.~Gupta and Y.-S. Ong, ``Back to the roots: Multi-x evolutionary computation,'' \emph{Cognitive Computation}, vol.~11, no.~1, pp. 1--17, 2019.

\bibitem{rios2021multitask}
T.~Rios, B.~van Stein, T.~B{\"a}ck, B.~Sendhoff, and S.~Menzel, ``Multitask shape optimization using a 3-d point cloud autoencoder as unified representation,'' \emph{IEEE Transactions on Evolutionary Computation}, vol.~26, no.~2, pp. 206--217, 2021.

\bibitem{martindale1990clockwork}
C.~Martindale, \emph{The clockwork muse: The predictability of artistic change.}\hskip 1em plus 0.5em minus 0.4em\relax Basic Books, 1990.

\bibitem{lengler2025runtime}
J.~Lengler, A.~Neumann, and F.~Neumann, ``Runtime analysis of evolutionary multitasking for classical benchmark problems,'' in \emph{Proceedings of the Genetic and Evolutionary Computation Conference}, 2025, pp. 1613--1621.

\bibitem{libreaking}
Y.~Li, J.~Liu, W.~Gong, Q.~Gu, Y.~Zhao, and Y.-S. Ong, ``Breaking multi-task curse: Reward-weighted evolution for black-box many-task optimization,'' in \emph{Proceedings of the 43rd International Conference on Machine Learning (ICML)}, 2026.

\bibitem{probst2020harmless}
M.~Probst and F.~Rothlauf, ``Harmless overfitting: Using denoising autoencoders in estimation of distribution algorithms,'' \emph{Journal of Machine Learning Research}, vol.~21, no.~78, pp. 1--31, 2020.

\bibitem{bengio2013generalized}
Y.~Bengio, L.~Yao, G.~Alain, and P.~Vincent, ``Generalized denoising auto-encoders as generative models,'' \emph{Advances in neural information processing systems}, vol.~26, 2013.

\bibitem{bhattacharjee2019estimation}
S.~Bhattacharjee and R.~Gras, ``Estimation of distribution using population queue based variational autoencoders,'' in \emph{2019 IEEE Congress on Evolutionary Computation (CEC)}.\hskip 1em plus 0.5em minus 0.4em\relax IEEE, 2019, pp. 1406--1414.

\bibitem{yu2022enhancing}
R.~Yu, Y.~Xu, H.~Peng, and M.-H. Chen, ``Enhancing population diversity by integrating iterative local search with deep convolutional generative adversarial networks (gans),'' in \emph{2022 26th International Conference on Pattern Recognition (ICPR)}.\hskip 1em plus 0.5em minus 0.4em\relax IEEE, 2022, pp. 4722--4728.

\bibitem{he2020evolutionary}
C.~He, S.~Huang, R.~Cheng, K.~C. Tan, and Y.~Jin, ``Evolutionary multiobjective optimization driven by generative adversarial networks (gans),'' \emph{IEEE transactions on cybernetics}, vol.~51, no.~6, pp. 3129--3142, 2020.

\bibitem{dang2024generative}
Q.~Dang, G.~Zhang, L.~Wang, S.~Yang, and T.~Zhan, ``A generative adversarial networks model based evolutionary algorithm for multimodal multi-objective optimization,'' \emph{IEEE Transactions on Emerging Topics in Computational Intelligence}, 2024.

\bibitem{zhang2025diffusion}
Y.~Zhang, B.~Hartl, H.~Hazan, and M.~Levin, ``Diffusion models are evolutionary algorithms,'' in \emph{International Conference on Learning Representations}, vol. 2025, 2025, pp. 74\,873--74\,894.

\bibitem{yan2024emodm}
X.~Yan and Y.~Jin, ``Emodm: A diffusion model for evolutionary multi-objective optimization,'' \emph{arXiv preprint arXiv:2401.15931}, 2024.

\bibitem{hartl2026heuristically}
B.~Hartl, Y.~Zhang, H.~Hazan, and M.~Levin, ``Heuristically adaptive diffusion-model evolutionary strategy,'' \emph{Advanced Science}, vol.~13, no.~20, p. e11537, 2026.

\bibitem{hansen2001completely}
N.~Hansen and A.~Ostermeier, ``Completely derandomized self-adaptation in evolution strategies,'' \emph{Evolutionary computation}, vol.~9, no.~2, pp. 159--195, 2001.

\bibitem{ul2020novel}
E.~ul~Haq, I.~Ahmad, and I.~M. Almanjahie, ``A novel parent centric crossover with the log-logistic probabilistic approach using multimodal test problems for real-coded genetic algorithms,'' \emph{Mathematical Problems in Engineering}, vol. 2020, no.~1, p. 2874528, 2020.

\bibitem{Deb1995RealcodedGA}
K.~Deb and A.~Kumar, ``Real-coded genetic algorithms with simulated binary crossover: Studies on multimodal and multiobjective problems,'' \emph{Complex systems}, vol.~9, no.~6, 1995.

\bibitem{deb2007self}
K.~Deb, K.~Sindhya, and T.~Okabe, ``Self-adaptive simulated binary crossover for real-parameter optimization,'' in \emph{Proceedings of the 9th annual conference on genetic and evolutionary computation}, 2007, pp. 1187--1194.

\bibitem{bali2019multifactorial}
K.~K. Bali, Y.-S. Ong, A.~Gupta, and P.~S. Tan, ``Multifactorial evolutionary algorithm with online transfer parameter estimation: Mfea-ii,'' \emph{IEEE Transactions on Evolutionary Computation}, vol.~24, no.~1, pp. 69--83, 2019.

\bibitem{gupta2022guest}
A.~Gupta, Y.-S. Ong, K.~A. De~Jong, and M.~Zhang, ``Guest editorial special issue on multitask evolutionary computation,'' \emph{IEEE Transactions on Evolutionary Computation}, vol.~26, no.~2, pp. 202--205, 2022.

\bibitem{gupta2016multiobjective}
A.~Gupta, Y.-S. Ong, L.~Feng, and K.~C. Tan, ``Multiobjective multifactorial optimization in evolutionary multitasking,'' \emph{IEEE transactions on cybernetics}, vol.~47, no.~7, pp. 1652--1665, 2016.

\bibitem{gupta2022half}
A.~Gupta, L.~Zhou, Y.-S. Ong, Z.~Chen, and Y.~Hou, ``Half a dozen real-world applications of evolutionary multitasking, and more,'' \emph{IEEE Computational Intelligence Magazine}, vol.~17, no.~2, pp. 49--66, 2022.

\bibitem{lim2021non}
R.~Lim, A.~Gupta, Y.-S. Ong, L.~Feng, and A.~N. Zhang, ``Non-linear domain adaptation in transfer evolutionary optimization,'' \emph{Cognitive Computation}, vol.~13, pp. 290--307, 2021.

\bibitem{preuss2005counteracting}
M.~Preuss, L.~Sch{\"o}nemann, and M.~Emmerich, ``Counteracting genetic drift and disruptive recombination in ($\mu$pluskomma$\lambda$)-ea on multimodal fitness landscapes,'' in \emph{Proceedings of the 7th annual conference on Genetic and evolutionary computation}, 2005, pp. 865--872.

\bibitem{syswerda1989uniform}
G.~Syswerda, ``Uniform crossover in genetic algorithms.'' in \emph{International Conference on Genetic Algorithms}, vol.~3, no. 2--9, 1989.

\bibitem{eiben1994genetic}
A.~E. Eiben, P.-E. Raue, and Z.~Ruttkay, ``Genetic algorithms with multi-parent recombination,'' in \emph{International conference on parallel problem solving from nature}.\hskip 1em plus 0.5em minus 0.4em\relax Springer, 1994, pp. 78--87.

\bibitem{wang2010study}
Y.~Wang, Z.~L{\"u}, and J.-K. Hao, ``A study of multi-parent crossover operators in a memetic algorithm,'' in \emph{International Conference on Parallel Problem Solving from Nature}.\hskip 1em plus 0.5em minus 0.4em\relax Springer, 2010, pp. 556--565.

\bibitem{kant2024identifiability}
M.~Kant, E.~Y. Ma, A.~Staicu, L.~J. Schulman, and S.~Gordon, ``Identifiability of product of experts models,'' in \emph{International Conference on Artificial Intelligence and Statistics}.\hskip 1em plus 0.5em minus 0.4em\relax PMLR, 2024, pp. 4492--4500.

\bibitem{gales2006product}
M.~J. Gales and S.~Airey, ``Product of gaussians for speech recognition,'' \emph{Computer Speech \& Language}, vol.~20, no.~1, pp. 22--40, 2006.

\bibitem{wong2025llm2tea}
M.~Wong, J.~Liu, T.~Rios, S.~Menzel, and Y.-S. Ong, ``Llm2tea: An agentic ai designer for discovery with generative evolutionary multitasking,'' \emph{IEEE Computational Intelligence Magazine}, vol.~20, no.~4, pp. 42--55, 2025.

\bibitem{jun2023shap}
H.~Jun and A.~Nichol, ``Shap-e: Generating conditional 3d implicit functions,'' \emph{arXiv preprint arXiv:2305.02463}, 2023.

\bibitem{weller1998tensorial}
H.~G. Weller, G.~Tabor, H.~Jasak, and C.~Fureby, ``A tensorial approach to computational continuum mechanics using object-oriented techniques,'' \emph{Computers in physics}, vol.~12, no.~6, pp. 620--631, 1998.

\bibitem{lin1991divergence}
J.~Lin, ``Divergence measures based on the shannon entropy,'' \emph{IEEE Transactions on Information theory}, vol.~37, no.~1, pp. 145--151, 1991.

\bibitem{kullback1951information}
S.~Kullback and R.~A. Leibler, ``On information and sufficiency,'' \emph{The annals of mathematical statistics}, vol.~22, no.~1, pp. 79--86, 1951.

\end{thebibliography}

\ifCLASSOPTIONcaptionsoff
  \newpage
\fi
\end{document}